\documentclass[letterpaper]{article} %
\usepackage{aaai2026}  %
\usepackage{times}  %
\usepackage{helvet}  %
\usepackage{courier}  %
\usepackage[hyphens]{url}  %
\usepackage{graphicx} %
\usepackage{natbib}  %
\usepackage{caption} %
\usepackage{algorithm}
\usepackage{algorithmic}

\newcommand{\norm}[1]{\left \lVert #1 \right \rVert}

\usepackage{url}            %
\usepackage{booktabs}       %
\usepackage{amsfonts}       %
\usepackage{nicefrac}       %
\usepackage{microtype}      %
\usepackage{xcolor}         %

\usepackage{bm}
\usepackage{mathtools}

\usepackage{newfloat}
\usepackage{listings}
\DeclareCaptionStyle{ruled}{labelfont=normalfont,labelsep=colon,strut=off} %
\floatstyle{ruled}
\newfloat{listing}{tb}{lst}{}
\floatname{listing}{Listing}
\title{DIP-GS: Deep Image Prior For Gaussian Splatting Sparse View Recovery}
\nocopyright
\author{
    Rajaei Khatib and
    Raja Giryes
}

\affiliations{
    Tel Aviv University\\
    rajaeikhatib@mail.tau.ac.il and raja@tauex.tau.ac.il
}

\usepackage{bibentry}

\begin{document}

\maketitle

\begin{abstract}
3D Gaussian Splatting (3DGS) is a leading 3D scene reconstruction method, obtaining high-quality reconstruction with real-time rendering runtime performance.
The main idea behind 3DGS is to represent the scene as a collection of 3D gaussians, while learning their parameters to fit the given views of the scene. While achieving superior performance in the presence of many views, 3DGS struggles with sparse view reconstruction, where the input views are sparse and do not fully cover the scene and have low overlaps. In this paper, we propose DIP-GS, a Deep Image Prior (DIP) 3DGS representation. By using the DIP prior, which utilizes internal structure and patterns, with coarse-to-fine manner, DIP-based 3DGS can operate in scenarios where vanilla 3DGS fails, such as sparse view recovery. Note that our approach does not use any pre-trained models such as generative models and depth estimation, but rather relies only on the input frames. Among such methods, DIP-GS obtains state-of-the-art (SOTA) competitive results on various sparse-view reconstruction tasks, demonstrating its capabilities.

\end{abstract}

\begin{figure}[ht!]
  \centering
    \includegraphics[width=0.52\textwidth]{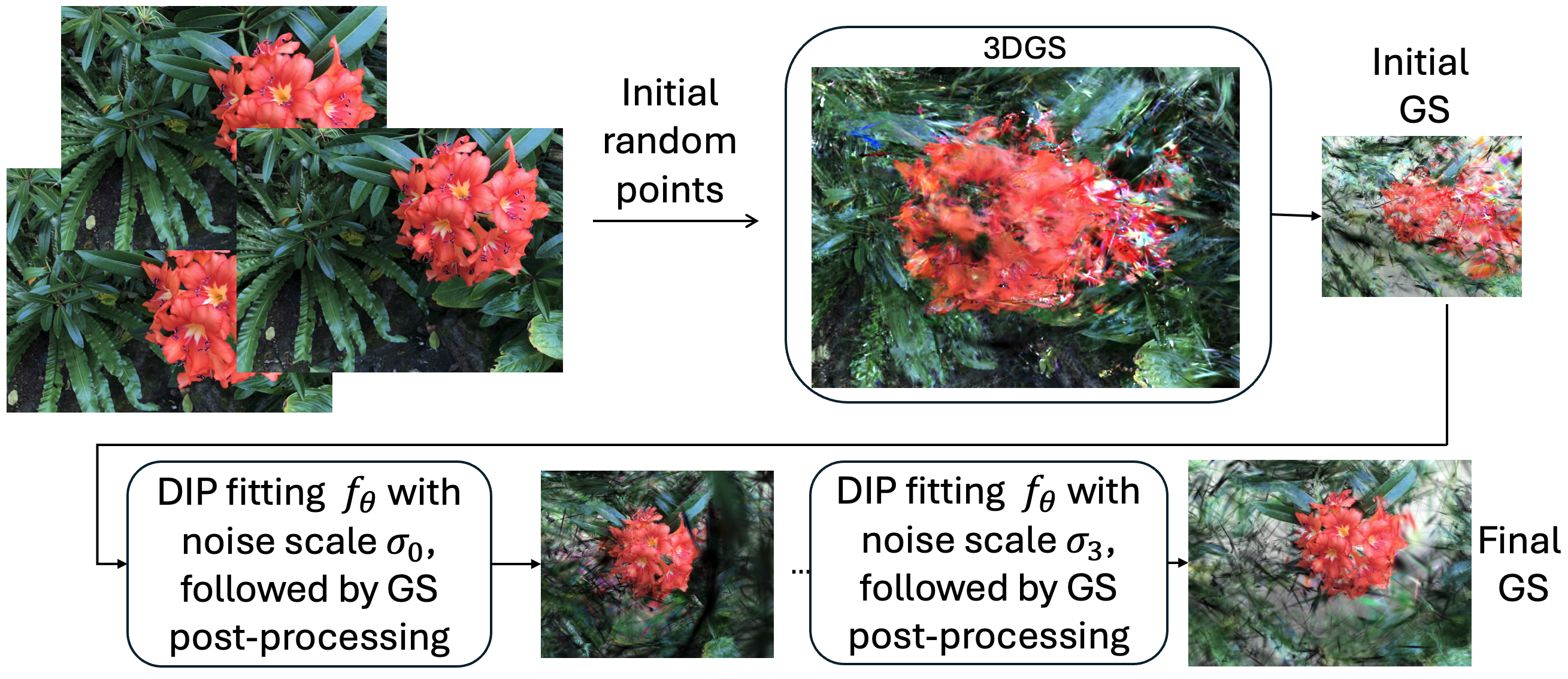}
    \caption{DIP-GS general scheme: First, the method starts by running vanilla 3DGS to get initial Gaussians. Next, DIP fitting and post-processing are applied sequentially.}
    \label{figure:general}
    \vspace{-0.2in}
\end{figure}

\section{Introduction}
\label{sec:intro}

The advent of 3D Gaussian Splatting (3DGS)~\cite{kerbl20233dsplatting} has revolutionized 3D scene representation, enabling high-quality reconstructions with unprecedented real-time rendering performance. This efficiency has surpassed many prior methods that, despite achieving quality, often struggled with high computational costs, large memory footprints, or low frame rates~\cite{mildenhall2021nerf,barron2021mip,barron2022mip,muller2022instant,khatib2024trinerflet}. Consequently, 3DGS has rapidly become a foundational technology for diverse real-time applications, including AR/VR, 3D generation, and interactive 3D editing, etc. \cite{tang2023dreamgaussian,ren2023dreamgaussian4d,qin2024langsplat,liu20243dgs,shen2024supergaussian,kerbl2024hierarchical,wu2024gaussctrl}.
At its core, 3DGS models a scene as a collection of 3D Gaussians, whose parameters are optimized to match a set of input views.

The main concept of 3DGS is to represent the scene as a group of 3D Gaussians, where each one is characterized by its center, opacity, scaling, rotation, and color. To render these Gaussians on a specific view, \citet{kerbl20233dsplatting} proposed an efficient differentiable rendering scheme as well. So, given a collection of input views of a scene with their known camera parameters, the Gaussians features are learned in an end-to-end optimization process, while pushing the rendered views to be as close as possible to the ground truth ones.

Despite its success with dense input views, vanilla 3DGS encounters significant challenges in sparse-view reconstruction. When input views are few and offer limited overlap, the optimization problem becomes severely ill-posed. This often leads to over-parameterization, where numerous erroneous Gaussian configurations can fit the sparse input views, resulting in poor generalization to novel viewpoints and unconvincing 3D geometry.
Addressing this sparsity challenge typically involves two main strategies. The first relies on external, pre-trained models to provide regularization, such as depth estimators or generative priors~\cite{li2024dngaussian,guedon2024matcha,zhu2024fsgs,fan2024instantsplat,liu2024reconx,sun2024dimensionx,wang2023sparsenerf}. While often effective, these methods are inherently limited by the diversity, robustness, and potential biases of the pre-trained models, and their performance can degrade if the pre-trained knowledge does not align well with the target scene. The second strategy aims to impose structural regularization directly on the 3DGS representation itself~\cite{yin2024fewviewgs,yang2023freenerf}. However, the typically unstructured and unordered nature of 3D Gaussians makes it non-trivial to design and enforce meaningful structural priors.

In this paper, we introduce DIP-GS, a novel approach that uniquely leverages DIP~\cite{ulyanov2018deep} to instill robust structural regularization for sparse-view 3DGS, \textit{without relying on any pre-trained models}. Our central novelty lies in re-purposing the inherent structure-inducing capabilities of DIP's convolutional neural network (CNN) architecture to directly generate a \textit{structured 2D grid of 3D Gaussian parameters}.
DIP was originally proposed for image restoration tasks, demonstrating that a randomly initialized CNN can capture low-level image statistics and internal patterns from a single degraded image to perform tasks like denoising or inpainting, effectively using the network structure itself as a prior. Inspired by this, and by observations that 3D Gaussian parameters can exhibit 'natural' structure when organized into a 2D grid~\cite{morgenstern2024compact}, DIP-GS makes two key conceptual leaps:

\textbf{1. Structured Gaussian Generation via DIP:} Instead of directly optimizing individual, unstructured Gaussians, DIP-GS learns a DIP network that maps a fixed random noise input to a 2D grid. Each "pixel" in this grid's output channels corresponds to the parameters (mean, scale, rotation, opacity, spherical harmonics) of a 3D Gaussian. This fundamentally changes the representation from an unstructured set to an implicitly structured one.

\textbf{2. Learning Network Weights as an Implicit Prior:} The optimization target shifts from the vast number of individual Gaussian parameters to the comparatively constrained weights of the DIP network. The CNN architecture itself, with its spatial biases, enforces a strong prior for local coherence and plausible global structure across the generated Gaussians. This inherent regularization is learned \textit{solely from the internal statistics of the sparse input views}.

This structured generation process, driven by optimizing the DIP network, provides a powerful inductive bias that helps to disambiguate scene geometry and appearance from limited data, mitigating the overfitting issues common in sparse-view 3DGS. Furthermore, we integrate this with a coarse-to-fine optimization scheme, allowing DIP-GS to progressively refine the scene details.

By having a strong internally derived prior, DIP-GS can operate effectively in scenarios where vanilla 3DGS fails and achieves state-of-the-art (SOTA) competitive results among pre-training free methods on various sparse-view reconstruction benchmarks. Our work demonstrates a novel pathway for robust 3DGS reconstruction by leveraging the implicit structural biases of neural network architectures, rather than explicit pre-training on external data.

\section{Related Works and Preliminaries}
\label{sec:preliminaries}

\textbf{3D Gaussian Splatting (3DGS)} \cite{kerbl20233dsplatting} is a 3D reconstruction approach, in which the scene is explicitly represented by a group of optimizable Gaussians. Each Gaussian is characterized by its mean $\bm{\mu} \in \mathbb{R}^{3}$, scale $\bm{s} \in \mathbb{R}^{3}$, rotation matrix $\bm{R} \in \mathbb{R}^{3 \times 3}$, opacity $o \in \mathbb{R}$, and spherical harmonics color coefficients $\bm{sh} \in \mathbb{R}^{3(l+1)^2}$ with degree $l$ for view-dependent color. The covariance matrix of each Gaussian is $\bm{\Sigma}=\bm{RSS}^T\bm{R}^T$, in which $\bm{S}$ is a diagonal matrix with the scaling $\bm{s}$ as the diagonal, thus the Gaussian is defined by $G(\bm{x})=e^{{-\frac{1}{2}}\bm{x}^T  \bm{\Sigma}^{-1} \bm{x}}$. Given view transform $\bm{W}$ and the Jacobian of the affine approximation of the projective
transformation $\bm{J}$ the covariance of the view projected (splatted) 2D Gaussian is $\bm{\Sigma}'=\bm{JW\Sigma W}^T\bm{J}^T$. To render the Gaussians, first they are sorted with respect to their distance from the view origin, then the color of each pixel $\hat{c}$ is calculated:

\begin{eqnarray}\label{eq:RENDER}
\begin{aligned}
\hat{c} = \sum_{i=1}^N T_i \cdot \alpha_i \cdot {c}_i, 
\quad \textrm{where}  \quad T_i = \prod_{j=1}^{i-1} (1-\alpha_i) .
\end{aligned}
\end{eqnarray}
where $c_i$ is the color of Gaussian $i$ in order, and $\alpha_i = o_i \cdot e^{{-\frac{1}{2}}\bm{\Delta}_i^T  \bm{\Sigma}'^{-1} \bm{\Delta}_i}$, where $\bm{\Delta}_i$ is the offset between the 2D mean of a splatted Gaussian and the pixel coordinate. The Gaussian parameters are optimized to minimize the reconstruction loss $\norm{c-\hat{c}}_2$, where $c$ is the pixel ground-truth color. The optimized rendering algorithm enables fast training and real-time FPS rendering of 3DGS. During training, Gaussians are added and removed via heuristics as in \citet{kerbl20233dsplatting,kheradmand20243d}.

\textbf{DIP} \cite{ulyanov2018deep}, given a degraded image $\bm{x}_0$, the method recovers its cleaned version $\hat{\bm{x}}$ by optimizing the parameters of a neural function $f_{\theta}$ that maps a uniform random noise image $\bm{z}$ to $\bm{x}_0$ by solving $min_{\theta} \norm{f_{\theta} (\tilde{\bm{z}})-\bm{x}_0}_2$, where $\tilde{\bm{z}} = \bm{z} + \sigma \bm{u}, \bm{u} \sim N(0,1)$ and $\sigma$ is a scale that is dependent of the recovery task and it prevents overfitting the degraded image. As demonstrated in \cite{ulyanov2018deep}, by using the prior bias of a neural network, DIP manages to understand the internal structure and patterns, and to recover natural images with respectable performance in tasks such as denoising, super-resolution, inpainting and etc, without any pre-training.

\textbf{3DGS Grid} \cite{morgenstern2024compact} is a scheme for organizing Gaussians in a 2d grid such that the features of the Gaussians form 2D grid channels with 'natural' structure. Thus, these channels can be compressed and saved using any image compression method, such as JPEG compression. Thus, by mainly re-organizing the Gaussians, this method managed to integrate image-based compression methods into 3DGS. Inspired by the same idea, we use DIP, which outputs a naturally structured 2D image, to represent the Gaussians.

\textbf{Sparse-View 3D Gaussian Splatting.}
While 3DGS~\cite{kerbl20233dsplatting} excels with dense input views, its performance significantly degrades in sparse-view scenarios due to the ill-posed nature of the task. This often leads to overfitting to the input views and poor novel view generalization. Several strategies have been proposed to mitigate this.

One common approach involves incorporating external priors, often derived from pre-trained models.
For instance, DNGaussian~\cite{li2024dngaussian} optimizes sparse-view 3D Gaussian radiance fields by introducing global-local depth normalization, leveraging depth information (potentially from monocular depth estimators) to regularize the geometry.
FSGS~\cite{zhu2024fsgs} focuses on real-time few-shot view synthesis using Gaussian Splatting, also implicitly or explicitly relying on strong priors learned during their specialized training or initialization to densify the sparse Gaussians effectively for novel view synthesis.
Similarly, FewViewGS~\cite{yin2024fewviewgs} employs view matching techniques, often using pre-trained 2D feature extractors, and a multi-stage training strategy to enhance reconstruction from few views. 
While effective, these methods can be constrained by the generalization capabilities of the pre-trained models they depend on (e.g., for depth estimation or feature extraction), or may require specific types of prior information that might not always be available or perfectly align with the target scene, which can influence the reconstruction quality.

Another line of work focuses on imposing structural regularization directly or learning priors from the input data itself, without relying on extensively pre-trained networks on external datasets.
For example, FreeNeRF~\cite{yang2023freenerf}, although developed for NeRFs, demonstrated improvements in few-shot neural rendering using frequency regularization, a self-driven prior that discourages high-frequency artifacts common in undertrained models.
However, imposing meaningful structural priors directly onto the typically unordered and unstructured set of 3D Gaussians can be challenging, as noted in the introduction.

Our proposed DIP-GS method aligns with this latter category but offers a distinct approach tailored for 3DGS. By leveraging DIP~\cite{ulyanov2018deep}, DIP-GS learns to generate a \textit{structured} 2D grid of Gaussian parameters from a random noise input. This inherently regularizes the Gaussian representation through the architectural biases of the CNN used in DIP, encouraging spatial coherence and plausible scene structure.
Unlike methods reliant on pre-trained depth estimators or 2D feature extractors, DIP-GS capitalizes on the internal statistics and self-similarity, guided by the structure-inducing capabilities of the DIP. This makes our approach pre-training free and more adaptable to diverse scenes where external priors might fail, introduce bias, or are simply unavailable. Our method's advantage lies in its ability to extract and enforce regularity from the sparse data alone, offering robust reconstruction by design without dependence on external pre-trained knowledge.

\section{DIP Gaussian Splatting (DIP-GS)}
\label{sec:method}

In DIP-GS, instead of learning the Gaussians features directly, we learn a DIP function $f_{\theta}$ that maps a uniform random noise image $\bm{z}$ to $\bm{x}_0$, forming an organized Gaussian grid. By using the prior bias of a neural network, we obtain a regularized structure for the Gaussians. 
In case of $n^2$ output Gaussians and a uniform random noise vector $\bm{z}$, the output Gaussian features are obtained by $(\bm{\mu},\bm{o},\bm{s},\bm{r},\bm{sh}) = f_{\theta}(\bm{z})$, where $\bm{\mu} \in \mathbb{R}^{n \times n \times 3}$, $\bm{o} \in \mathbb{R}^{n \times n \times 1}$, $\bm{s} \in \mathbb{R}^{n \times n \times 3}$, $\bm{r} \in \mathbb{R}^{n \times n \times 4}$, $\bm{sh} \in \mathbb{R}^{n \times n \times 3(l+1)^2}$ with degree $l$ for view-dependent color. 
As shown in \cite{kerbl20233dsplatting}, the rotation matrices $\bm{R}$ are build from the quaternion vector $\bm{r}$ and scale matrices $\bm{S}$ contain $\bm{s}$ in the diagonal. 
After obtaining the 2D organized Gaussian grid as the output of $f_{\theta}$, the grid is splitted to form the desired Gaussian structure, where each entry in the grid corresponds to a single Gaussian, and the Gaussian features are the corresponding parameters of this entry, resulting in $n^2$ Gaussians $\{ (\bm{\mu}_{i,j},\bm{o}_{i,j},\bm{s}_{i,j},\bm{r}_{i,j},\bm{sh}_{i,j})\}_{0 \leq i,j < n }$, as in \cite{morgenstern2024compact}. 
It is worth mentioning that the Gaussian 2D organized grid is only an internal structure, which is split to unordered collection to form a regular 3DGS structure, and from an outside black-box perspective, this method generates regular unordered set of Gaussians.
Thus, these Gaussians can be passed to any Gaussian related pipeline, such as rendering. The function $f_{\theta}$ is compound of several DIP networks, where each one of the $5$ channels is regressed by a separate DIP, however, they all operate on the same input, $f_{\theta}(\bm{z}) = (f_{\theta_\mu}^{\mu}(\bm{z}),f_{\theta_o}^{o}(\bm{z}),f_{\theta_s}^{s}(\bm{z}),f_{\theta_r}^{r}(\bm{z}),f_{\theta_{sh}}^{sh}(\bm{z}))$, in which they output the Gaussian features $(\bm{\mu},\bm{o},\bm{s},\bm{r},\bm{sh})$.

\begin{figure*}[ht]
    \centering
    \includegraphics[width=\textwidth]{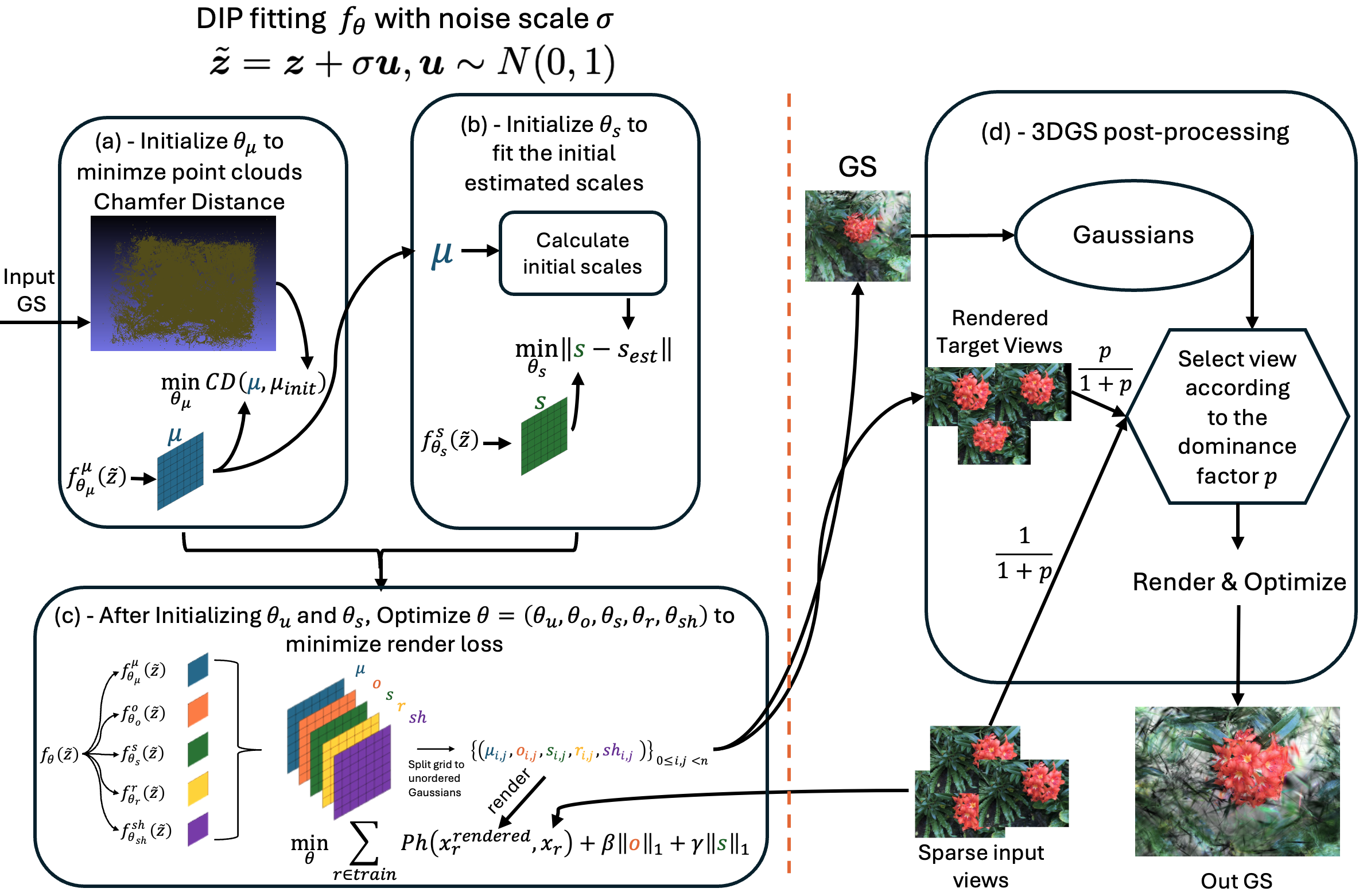}
    \caption{DIP-GS components at a given noise level. (a) - First, the mean's network $f_{\theta_{\mu}}^{\mu}$ is initialized by minimizing the point cloud Chamfer Distance between its output $\bm{\mu}$, which is mapped from the noise $\tilde{\bm{z}}$, and the initial Gaussians means $\bm{\mu}_{init}$. (b) - Second, the scale's network $f_{\theta_{s}}^{s}$ is initialized by fitting the output scale channel $\bm{s}$, which is mapped from the noise $\tilde{\bm{z}}$, to the estimated scale guess $\bm{s}_{est}$. (c) - Next, the DIP optimization, in which $f_{\theta}$ maps $\tilde{\bm{z}}$ to the Gaussian features, and $\theta$ is learned by minimizing the render loss alongside other regularizations. (d) - The post-processing stage, where the Gaussians are initialized by the output of the DIP $f_{\theta}$ that was trained in the previous stage. At each step, the method chooses a frame either from the sparse input views or one of the target views.}
    \label{figure:method}
    \vspace{-0.2in}
\end{figure*}

Given a sparse collection of views of a given scene, first, we start by running the vanilla 3DGS on them with random initialization to obtain an initial Gaussians estimation $(\bm{\mu}_{init},\bm{o}_{init},\bm{s}_{init},\bm{r}_{init},\bm{sh}_{init})$, as shown in Figure \ref{figure:general}. As shown in \cite{yin2024fewviewgs,li2024dngaussian} and Figure \ref{figure:general}, the initial estimation usually describes inaccurate geometry with poor novel view performance. DIP-GS starts by fitting the Gaussian centers $\bm{\mu}=f_{\theta_\mu}^{\mu}(\bm{z})$ to be close to the initial estimation centers $\bm{\mu}_{init}$ by minimizing the Chamfer Distance (CD) between these two point clouds:
\begin{eqnarray}\label{eq:init}
\begin{aligned}
\min_{\theta_{\mu}} \,  CD(f_{\theta_{\mu}}^{\mu}(\tilde{\bm{z}}), \bm{\mu}_{init}) 
\end{aligned}
\end{eqnarray}
in which $\tilde{\bm{z}} = \bm{z} + \sigma \bm{u}, \bm{u} \sim N(0,1)$, and $\sigma$ is the noise scale. The operation is illustrated in Figure \ref{figure:method}-(a).

After initializing the Gaussian means, the Gaussian scales $\bm{s} = f_{\theta_s}^{s}(\bm{z})$ are initialized. To do that, an initial guess of the scale of each Gaussian $\bm{s}_{est}$ is calculated to be the average distance of the $3$ closest Gaussians from the previously initialized means $\bm{\mu}$, as in vanilla 3DGS. Next, the Gaussian scales $\bm{s}=f_{\theta_s}^{s}(\bm{z})$ are fitted to be as close as possible to the initial scale guess $\bm{s}_{est}$ (Figure \ref{figure:method}-(b)):
\begin{eqnarray}\label{eq:init_scale}
\begin{aligned}
\min_{\theta_s} \,  \norm{f_{\theta_s}^s(\tilde{\bm{z}})-\bm{s}_{est}}
\end{aligned}
\end{eqnarray}

This initialization process is important for the method's convergence, since it is known in the GS literature that a good initialization is crucial for good recovery, as is demonstrated in the experiment section. Next, the rendering-based optimization process starts, where the DIP parameters are learned to minimize the rendered view loss:
\begin{eqnarray}\label{eq:DIPGS_opt}
\begin{aligned}
\min_{\theta} \,  \sum_{r \in train} Ph(\pi_r (f_{\theta}(\tilde{\bm{z}})), \bm{x}_r)  + \\ \beta \norm{(f_{\theta_o}^o(\tilde{\bm{z}})}_1 +  \gamma \norm{(f_{\theta_s}^s(\tilde{\bm{z}})}_1
\end{aligned}
\end{eqnarray}
where $Ph$ is the rendering photometric loss used in \cite{kerbl20233dsplatting}, $\pi_r$ is the Gaussian rendering operation, $\bm{x}_r$ the ground-truth image, and $\beta,\gamma$ are parameters for the opacity and scale regularization, see Figure \ref{figure:method}-(c).

Since scene recovery from sparse views is an ill-posed problem, in some cases, floaters close to the camera may appear in the novel views, as shown in \cite{yang2023freenerf}. Thus, to prevent that, an occlusion regularization is added to penalize Gaussians that are close to the cameras, in the same spirit of the occlusion regularization introduced in FreeNeRF \cite{yang2023freenerf}. Given a Gaussian and a view, the bounding box corners of the Gaussian are estimated, then the depth of the closest corner to the view origin is calculated - $d$, and the occlusion regularization for this Gaussian on this view is $o \cdot ReLU(1-d/d_{min})$, where $o$ is its opacity and $d_{min}$ is a hyperparameter. Thus, if the depth $d$ is less than $d_{min}$ (close to the view) the Gaussian is forced to either move away from the view or have low opacity. The final occlusion regularization $occ\_reg$ is obtained by calculating the mean of this term over all Gaussians with all views, and the hyperparameter $\delta$ as its weight in the loss term.

When the optimization finishes, the output Gaussian are obtained by using the original noise vector $\bm{z}$ (and not $\hat{\bm{z}}$) $(\bm{\mu},\bm{o},\bm{s},\bm{r},\bm{sh}) = f_{\theta}(\bm{z})$. Next, a post-processing stage is applied using vanilla GS with these Gaussians as the initial Gaussians, and their test views renderings for test views supervision. At each iteration, the method selects an image from the estimated test views with $\frac{p}{1+p}$ probability and an image from the input sparse views with $\frac{1}{1+p}$ probability, where $p$ is a hyperparameter called dominance factor. Since in the DIP optimization the Gaussians are not directly optimized, this means that Gaussians cannot be cloned or split, thus, the GS post-processing operation comes to introduce densification into this scheme. This operation is shown in Figure \ref{figure:method}-(d).

DIP-GS operates in a coarse-to-fine manner, it first starts with a high noise scale $\sigma$, since the initial Gaussians estimation is not accurate. Following that, the output Gaussians from the GS postprocess step will be the input Gaussian for the DIP-GS fit process again, followed by another GS post-process, but this time with equal or smaller noise scale. Thus, the DIP-GS fit then GS post process stages are applied several times with non-increasing noise level values, where the output Gaussians of each stage are the input for the next one. This coarse-to-fine approach enables a sequential recovery of the scene, where at the beginning, when the noise scale is high, the recovered scene will be blurry, but with good generalization in the unseen views. Then, when the noise level drops gradually, more and more details will be recovered. DIP-GS general scheme is described in Figure \ref{figure:general}.

\begin{figure}[ht]
    \centering
    \includegraphics[width=0.42\textwidth]{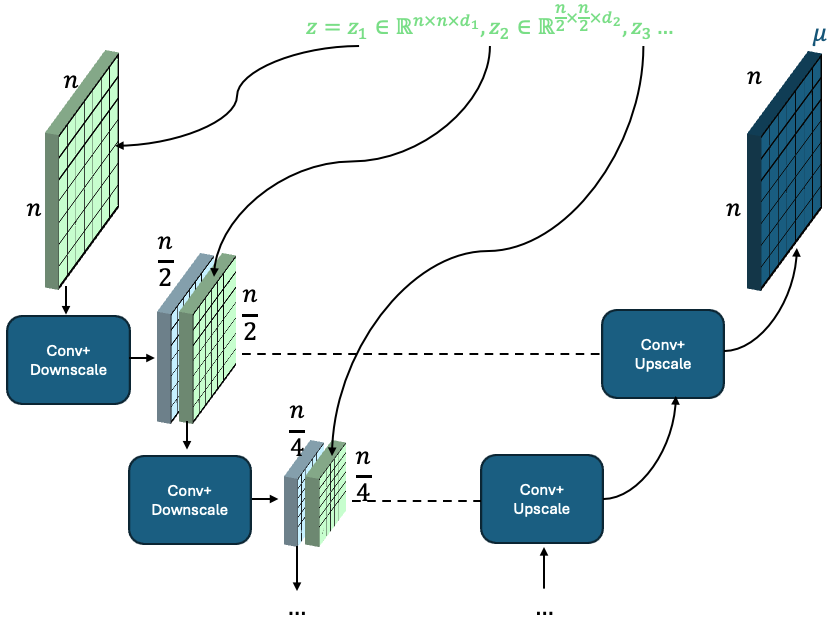}
    \caption{DIP Unet Architecture. The noise vector $\bm{z}$ (green) is fed as an input and also injected and concatenated into intermediate outputs (blue). Dash-lines are the skip cons.
    }
    \label{figure:unet}
    \vspace{-0.2in}
\end{figure}

The DIP functions are represented by an Encoder-Decoder Unet architecture with skip connections, where each layer is compound of a convolution followed by a normalization and then activation. Unlike in DIP, where the uniform random noise vector $\bm{z}$ was only fed to the Encoder as its input, in this method $\bm{z}$ is fed also to the inner intermediate layers. Thus, in case of $n^2$ output Gaussians, $\bm{z}=\bm{z}_1,\bm{z}_2,...$ in which $\bm{z}_1 \in \mathbb{R}^{n \times n \times d_1}$ is the input to the network (The equivalent of DIP), $\bm{z}_2 \in \mathbb{R}^{\frac{n}{2} \times \frac{n}{2} \times d_2}$ which will be concatenated to the output of the first downscaling layer in the network, and so on. The injection of the random vector into the intermediate results reduces the overfitting and ensures a plausible generalization output to the unseen areas, see Figure \ref{figure:unet}.

\begin{table}[ht]
\centering
\resizebox{0.48\textwidth}{!}{
    \begin{tabular}{ p{0.2\textwidth} | p{0.1\textwidth} | p{0.1\textwidth} | p{0.1\textwidth} }
      Method &  PSNR$\uparrow$ & SSIM$\uparrow$ & LPIPS$\downarrow$ \\
     \hline
     3DGS w/o occ reg & $14.452$ & $0.734$ & $0.223$ \\
     \hline
     3DGS w occ reg& $16.299$ & $0.744$ & $0.206$ \\
     \hline
     1st stage & $18.454$ & $0.797$ & $0.174$\\
     \hline
     2nd stage & $19.225$ & $0.816$ & $0.159$\\
     \hline
     3rd stage & $19.657$ & $0.828$ & $0.150$\\
     \hline
     \hline
     4th stage - final ours & $19.798$ & $0.836$ & $0.144$\\
     \hline
     \hline
     Unet with $\bm{z}$ as only input & $18.255$ & $0.808$ & $0.163$ \\
     \hline
     W/o opacity reg & $19.084$ & $0.813$ & $0.159$ \\
     \hline
     W/o scale reg & $19.477$ & $0.821$ & $0.154$ \\
    \hline
    W/o occ reg & $14.106$ & $0.783$ & $0.200$ \\
    \hline
     W/o $\bm{\mu}$ init & $15.484$ & $0.720$ & $0.233$ \\

     \hline
     W/o $\bm{s}$ init & $17.501$ & $0.785$ & $0.182$ \\

     \hline
     One Unet & $18.497$ & $0.792$ & $0.171$ \\

     \hline
     Two Unets & $19.070$ & $0.830$ & $0.1485$ \\

     \hline
     Dom. factor $p=0$. & $18.066$ & $0.804$ & $0.160$ \\

     \hline
     
    \end{tabular}}
    \caption{Ablation table on DTU to illustrate the contribution of DIP-GS design choices. In up-to-down order, first the contribution of the occ. reg on the inital stage. Then the contribution of the coarse-to-fine strategy. Next, the method is applied without Unet intermediate injection, opacity, scale and occ. regularizations, mean and scale initialization. Following that, the method is tested with one and two Unets. Finally, it is tested without dominance factor.}
    \label{tab:ablation}
     \vspace{-0.2in}
\end{table}

\begin{table*}[h]
\centering
\resizebox{\textwidth}{!}{
    \begin{tabular}{ p{0.12\textwidth} | p{0.10\textwidth}  p{0.10\textwidth} p{0.10\textwidth}
    |p{0.10\textwidth}  p{0.10\textwidth} p{0.10\textwidth}  
    |p{0.10\textwidth}  p{0.10\textwidth} p{0.10\textwidth}
    }
      &  \multicolumn{3}{c|}{Blender} & \multicolumn{3}{c}{LLFF} & \multicolumn{3}{c}{DTU} \\

      & \multicolumn{3}{c|}{$8$ Images} & \multicolumn{3}{c|}{$3$ Images} & \multicolumn{3}{c|}{$3$ Images} \\
      
     Method &  PSNR$\uparrow$  & SSIM$\uparrow$ & LPIPS$\downarrow$ &  PSNR$\uparrow$ & SSIM$\uparrow$ & LPIPS$\downarrow$ &  PSNR$\uparrow$ & SSIM$\uparrow$ & LPIPS$\downarrow$ \\
     \hline
     
     DietNeRF* &  $23.147$ & $0.866$ & $0.109$ &   $14.94$ & $0.370$ & $0.496$ & $11.85$ & $0.633$ & $0.314 $\\
     \hline
     FewViewGS* &  $\underline{25.550}$ & $0.886$ & $0.092$ &  ${18.96}$ & ${0.585}$ & ${0.307}$ & $19.13$ & $\underline{0.792} $ & $0.186$\\
     \hline
     \hline
     RegNeRF** &  $23.86$ & $0.852$ & $0.105$ &  $19.08$ & $0.587$ & $0.336$ & $18.89$ & $0.745$ & $0.190 $\\
     \hline
     
     SparseNeRF** &  $24.04$ & $0.876$ & $0.113$ &  $19.86$ & $0.624$ & $0.328$ & $19.55 $ & $0.769$ & $0.201$\\
     \hline
     
     DNGaussian** &  $24.305$ & $0.886$ & $\underline{0.088}$ &  $19.12$ & $0.591 $ & $0.294$
     & $18.91$ & $0.790$ & $\underline{0.176}$
     \\
     \hline
     FSGS** &  $24.64$ & $\underline{0.895}$ & $0.095$  &  $\textbf{20.31}$ & $\underline{0.652}$ & $\underline{0.288}$ & $-$ & $-$ & $-$\\
     \hline
     \hline
     Mip-NeRF &  $20.89$ & $0.830$ & $0.168$ & $14.62$ & $0.351$ & $0.495$ & $8.68$ & $0.571$ & $0.353$\\
     \hline
     3DGS & $22.226$ & $0.858$ & $0.114$ &  $15.52 $ & $0.408 $ & $0.405$ & $14.45$ & $0.734$ & $0.223$\\
     \hline
     FreeNeRF &  $24.259$ & $0.883$ & $0.098$ &  $19.63$ & $0.612$ & $0.308$ & $\textbf{19.92}$ & $0.787$ & $0.182$\\
     \hline
     \hline
     
    DIP-GS (ours) &  $\textbf{25.90}$ & $\textbf{0.898}$ & $\textbf{0.087}$ &  $\underline{20.13}$ & $\textbf{0.662}$ & $\textbf{0.221}$ & $\underline{19.79}$ & $\textbf{0.836}$ & $\textbf{0.144}$\\
     \hline
    \end{tabular}
}
    \caption{Sparse recovery on Blender, LLFF and DTU datasets. \textbf{Bold} is best, \underline{underline} is second. First category methods (no pre-trained used) are not denoted, methods from the second category (non 3D-aware pre-trained used) are denoted with *, and methods from the third category (3D-aware pre-trained used) are denoted with **. 
    For fair comparison with FewViewGS, we attach the results with random init for the initial 3DGS stage, such as in DIPGS and the other methods.}
    \label{tab:blender_llff_resutls}
     \vspace{-0.2in}
\end{table*}

\begin{table*}[h]
\centering
\resizebox{\textwidth}{!}{
    \begin{tabular}{ p{0.12\textwidth} | p{0.07\textwidth}  p{0.07\textwidth} p{0.07\textwidth}
    |p{0.07\textwidth}  p{0.07\textwidth} p{0.07\textwidth}  
    |p{0.07\textwidth}  p{0.07\textwidth} p{0.07\textwidth}
    |p{0.07\textwidth}  p{0.07\textwidth} p{0.07\textwidth}
    }
      &  \multicolumn{6}{c|}{LLFF}  & \multicolumn{6}{c}{DTU} \\

      & \multicolumn{3}{c|}{$6$ Images} & \multicolumn{3}{c|}{$9$ Images} & \multicolumn{3}{c|}{$6$ Images} & \multicolumn{3}{c|}{$9$ Images} \\
      
     Method &  PSNR$\uparrow$  & SSIM$\uparrow$ & LPIPS$\downarrow$ &  PSNR$\uparrow$ & SSIM$\uparrow$ & LPIPS$\downarrow$ &  PSNR$\uparrow$ & SSIM$\uparrow$ & LPIPS$\downarrow$ &  PSNR$\uparrow$ & SSIM$\uparrow$ & LPIPS$\downarrow$\\
     \hline
     
     FewViewGS* &  $21.33$ & $0.688$ & $0.220$ &  $23.09$ & $0.769 $ & $0.164$ & $\underline{23.51}$ & $\underline{0.891}$ & $0.123$ & $\underline{25.75}$ & $\underline{0.925}$ & $0.101$\\
     \hline
     \hline
     RegNeRF** &  $23.10$ & $0.760$ & $0.206$ &  $24.86$ & $0.820$ & $0.161$ & $19.10$ & $0.757$ & $0.233$ & $22.30$ & $0.823$ & $0.184$\\
     \hline

     DNGaussian** &  $22.18$ & $0.755$ & $0.198$ &  $23.17$ & $0.788$ & $0.180$ & $-$ & $-$ & $-$ & $-$ & $-$ & $-$
     \\
     \hline
     FSGS** &  $\textbf{24.55}$ & $\underline{0.795}$ & $\underline{0.177}$ &  $\textbf{25.89}$ & $\underline{0.845}$ & $\underline{0.143}$ & $-$ & $-$ & $-$ & $-$ & $-$ & $-$\\
     \hline
     \hline
     Mip-NeRF &  $22.91$ & $0.756$ & $0.213$ &  $24.88$ & $0.826$ & $0.160$ & $-$ & $-$ & $-$ & $-$ & $-$ & $-$\\
     \hline
     3DGS &  $20.36$ & $0.664$ & $0.252$ &  $21.49$ & $0.717$ & $0.254$ & $21.85$ & $0.870$ & $\underline{0.122}$  & $24.60$ & $0.918$ & $\textbf{0.086}$\\
     \hline
     FreeNeRF &  $23.73$ & $0.779$ & $0.195$ &  $25.13$ & $0.827$ & $0.160$ & $22.39$ & $0.779$ & $0.240$ & $24.20$ & $0.833$ & $0.187$\\
     \hline
     \hline
     
    DIP-GS (ours) &  $\underline{24.39}$ & $\textbf{0.829}$ & $\textbf{0.125}$ &  $\underline{25.57}$ & $\textbf{0.855}$ & $\textbf{0.108}$ & $\textbf{24.15}$ & $\textbf{0.902}$ & $\textbf{0.104}$ & $\textbf{25.97}$ & $\textbf{0.928}$ & $\underline{0.087}$\\
     \hline
    \end{tabular}
}
    \caption{$6$ and $9$ views scene recovery on LLFF and DTU datasets. \textbf{Bold} is best, \underline{underline} is second. As noticed, DIP-GS outperforms other methods in most of the cases. For fair comparison with FewViewGS, we attach the results with random init for the initial 3DGS stage, such as in DIP-GS and the other methods.}
    \label{tab:llff_dtu_6_9_views}
     \vspace{-0.2in}
\end{table*}

\section{Experiment Results}
\label{sec:exps}
In the sparse view recovery task, the goal is to recover the scene given sparse input views with the camera parameters. The proposed method is applied on several datasets, the first one is the Blender \cite{mildenhall2021nerf} dataset, which consists of 8 synthetic scenes with complex geometry and coloring. The second one is LLFF \cite{mildenhall2019local} dataset, which consists of 8 real-world scenes, and the third one is composed of $15$ selected scenes from the DTU \cite{jensen2014large} dataset. Following the setup in \cite{li2024dngaussian} in the Blender case, the method is trained on 8 specific images per scene, with $\frac{1}{2}$ of the original resolution. For LLFF and DTU, the method is trained on 3 specific images per scene with $\frac{1}{8}$ and $\frac{1}{4}$ of the original resolution respectively.

\noindent \textbf{Implementation details.} 
We provide below the implementation details of the various stages of our DIP-GS pipeline:

\textit{1. Initial 3DGS Estimation:}
The process commences by running a modified version of vanilla 3DGS. The primary modification is that the opacity reset operation commonly used in 3DGS densification is disabled. Also, an opacity $\beta_{init}$ and scaling $\gamma_{init}$ regularization term are introduced, as in \cite{kheradmand20243d}, alongside occlusion regularization $\delta_{init}$ for some cases. The initial 3D Gaussians are seeded from random points within the scene's bounding box. Let $N_{init}$ be the number of Gaussians obtained from this stage.

\textit{2. DIP-GS Network Architecture and Initialization:}
The core of our method, the DIP-GS generator $f_{\theta}$, is designed to produce $n^2 = 0.75 \times N_{init}$ Gaussians. This generator $f_{\theta}$ is comprised of five distinct U-Net architectures~\cite{ronneberger2015u} with $3$ down/up scale stages, each dedicated to regressing a specific component of the Gaussian parameters: mean $\bm{\mu} \in \mathbb{R}^{n \times n \times 3}$, opacity $o \in \mathbb{R}^{n \times n \times 1}$, scale $\bm{s} \in \mathbb{R}^{n \times n \times 3}$, rotation (represented as a quaternion $\bm{r} \in \mathbb{R}^{n \times n \times 4}$), and spherical harmonics coefficients $\bm{sh} \in \mathbb{R}^{n \times n \times 3}$ (where the SH degree is chosen $l=0$). All five U-Nets receive the same fixed 
random noise tensor $\bm{z}$ as input, where the input channel dimension $d$ is set to $32$, and the intermediate channels dimensions are $16,32,64$. 
Each element of this input noise tensor $\bm{z}$ is independently sampled from a uniform distribution $U(0, 0.1)$, as in DIP.

\textit{3. DIP-GS Optimization:}
The optimization of the DIP-GS network parameters $\theta$ proceeds in 3 phases: (i) Chamfer Distance Fitting (Means): For an initial $3000$ iterations, only the U-Net responsible for generating the Gaussian means ($\bm{\mu} = f_{\theta}^{\mu}(\bm{z})$) is optimized. The objective is to minimize the point cloud Chamfer Distance between these generated means and the means of the Gaussians obtained from the initial 3DGS estimation stage, see Figure \ref{figure:method}-(a); and (ii) Scale Fitting: For an initial $3000$ iterations, only the U-Net responsible for generating the Gaussian scales ($\bm{s} = f_{\theta}^{s}(\bm{z})$) is optimized to be as close as possible to the estimates scales, see Figure \ref{figure:method}-(b); and (iii) Rendering Loss Optimization (Full Network): Subsequently, for $4000$ iterations, the parameters $\theta$ of all five U-Nets are jointly optimized by minimizing a photometric rendering loss between the rendered views from DIP-GS and the ground truth input views. During this phase, the opacity, scale and occlusion regularizations are added with weights $\beta, \gamma$, and $\delta$ respectively.

\textit{4. Post-processing with Vanilla 3DGS:}
After the DIP-GS optimization, the Gaussians generated by $f_{\theta}(\bm{z})$ are used to initialize a standard vanilla 3DGS optimization. This post-processing step helps to further refine the Gaussians and introduce densification where needed. During this vanilla 3DGS optimization, a dominance factor $p=0.1$ is used for selecting between training on input views versus self-supervision from rendered test views. Similar to previous stages, opacity, scale and occlusion regularizations are also added with weights $\beta_{post}, \gamma_{post}$, and $\delta_{post}$, respectively. The values of these hyper-parameters are attached in supp. mat. It is worth emphasizing that scale occlusion regularizations are only used with DTU dataset.

\textit{5. Coarse-to-Fine Strategy:}
The entire procedure, encompassing DIP-GS optimization (steps 3a and 3b) followed by GS post-processing (step 4), is embedded within a coarse-to-fine framework spanning $4$ stages. In each stage $k$, the input noise tensor to the DIP U-Nets is perturbed as $\bm{z}'_k = \bm{z} + \sigma_k \cdot \mathcal{N}(0,I)$, where $\mathcal{N}(0,I)$ is standard Gaussian noise and $\sigma_k$ is a stage-dependent noise level. The sequence of noise levels used is $\bm{\sigma} = [\sigma_1, \sigma_2, \sigma_3, ...]$ (e.g., progressively non-increasing values such as $\bm{\sigma} = [0.0333,0.01,0.005,0.002]$ for our case. This allows the model to first capture coarse scene structure with higher noise and then refine details as the noise level decreases.

\noindent \textbf{Ablation Studies.}
We conduct ablation studies on the DTU dataset to assess the contribution of each component in DIP-GS, see Tab. \ref{tab:ablation}. The results, in up-to-down order, show that \emph{occlusion regularization} improves performance in the initial 3DGS stage, while the \emph{coarse-to-fine strategy} with decreasing noise levels enables the model to progressively refine details. \emph{Injecting the noise vector z into intermediate UNet layers} rather than only at the encoder input yields better results, and adding \emph{opacity, scale, and occlusion regularizations} provides a notable boost. Removing \emph{mean and scale initialization} significantly degrades performance, confirming its importance. Additionally, using \emph{5 specialized UNets} outperforms simpler configurations such as one UNet for all features or two UNets, one for the mean feature and the other for the rest. Finally, the \emph{dominance factor} in post-processing proves essential for the final performance.

\begin{figure*}[h]
\centering
    \begin{tabular}{   p{0.17\textwidth}  p{0.17\textwidth} p{0.17\textwidth}  p{0.17\textwidth} p{0.17\textwidth}}

    \includegraphics[width=0.19\textwidth]{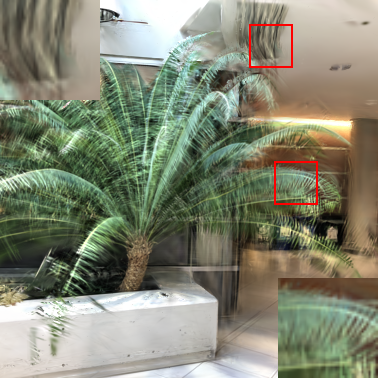} &
    \includegraphics[width=0.19\textwidth]{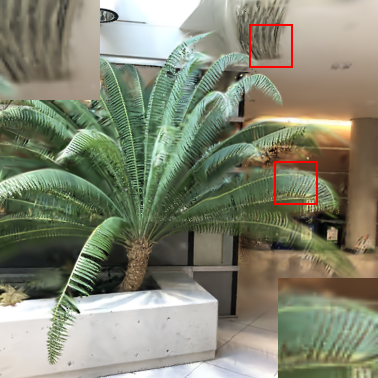} &
    \includegraphics[width=0.19\textwidth]{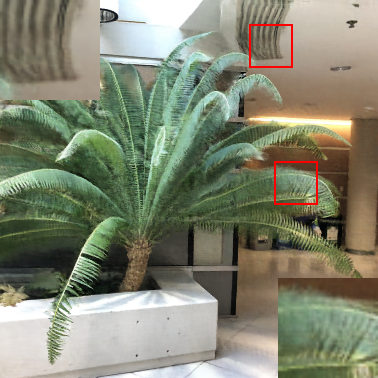} &    \includegraphics[width=0.19\textwidth]{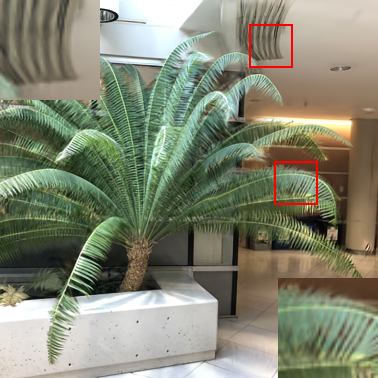} &
    \includegraphics[width=0.19\textwidth]{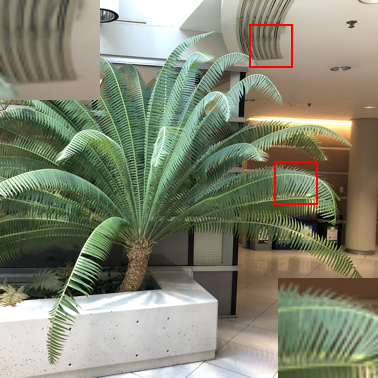}  \\

    \includegraphics[width=0.19\textwidth]{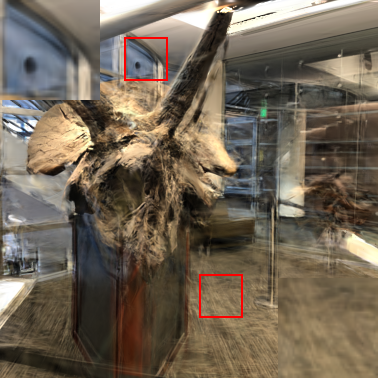} &
    \includegraphics[width=0.19\textwidth]{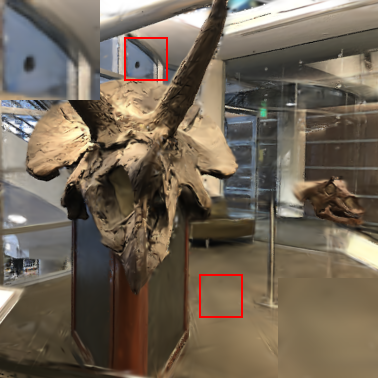} &
    \includegraphics[width=0.19\textwidth]{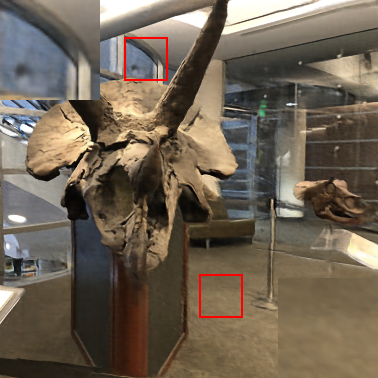} &    \includegraphics[width=0.19\textwidth]{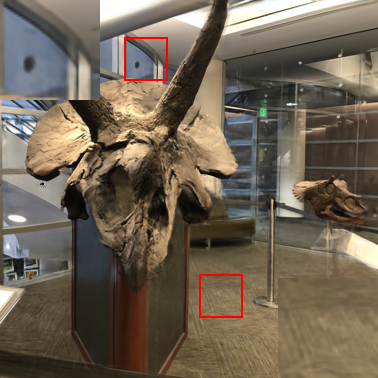} &
    \includegraphics[width=0.19\textwidth]{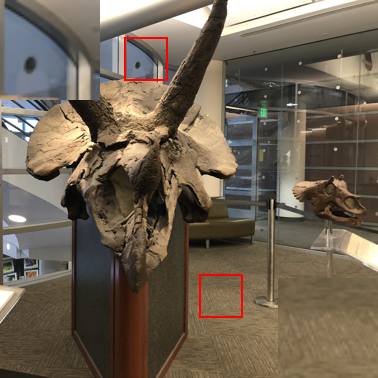}  \\

    \includegraphics[width=0.19\textwidth]{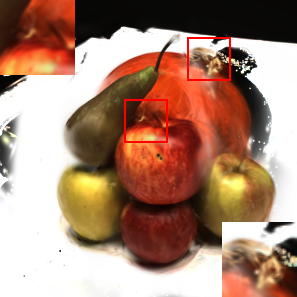} &
    \includegraphics[width=0.19\textwidth]{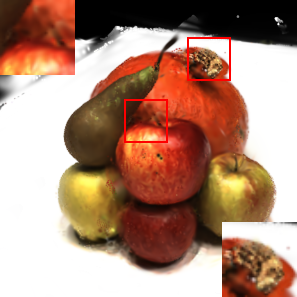} &
    \includegraphics[width=0.19\textwidth]{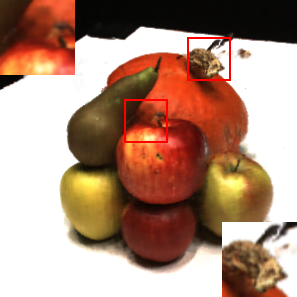} &    \includegraphics[width=0.19\textwidth]{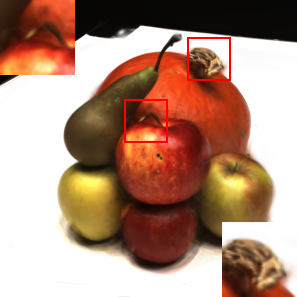} &
    \includegraphics[width=0.19\textwidth]{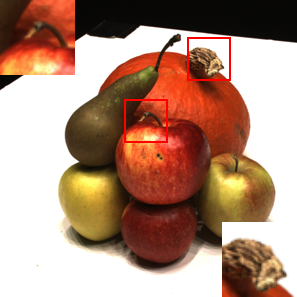}  \\

    \includegraphics[width=0.19\textwidth]{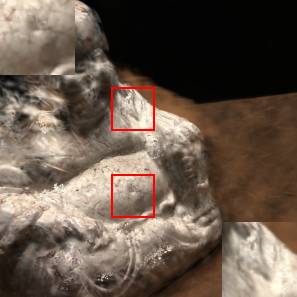} &
    \includegraphics[width=0.19\textwidth]{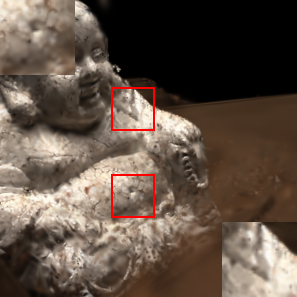} &
    \includegraphics[width=0.19\textwidth]{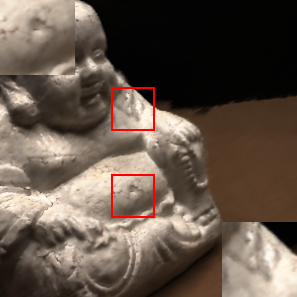} &    \includegraphics[width=0.19\textwidth]{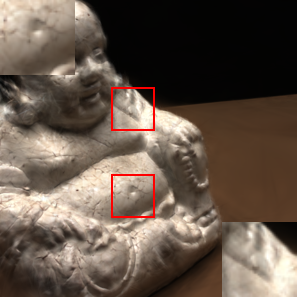} &
    \includegraphics[width=0.19\textwidth]{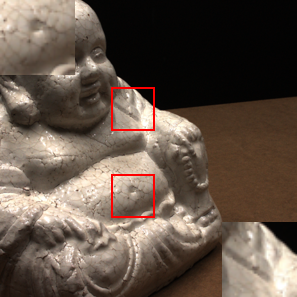}  \\

    \multicolumn{1}{c}{3DGS} & \multicolumn{1}{c}{DNGaussian}  & \multicolumn{1}{c}{FreeNeRF} &\multicolumn{1}{c}{DIP-GS} & \multicolumn{1}{c}{GT}

    \end{tabular}
    \caption{DTU and LLFF qualitative results. The qualitative results clearly illustrate the capabilities of DIP-GS.}
    \label{fig:qualitive_results_llff}
    \vspace{-0.2in}
\end{figure*}

\noindent \textbf{Comparison with State-of-the-Art.}
We compare DIP-GS with both NeRF and 3DGS based sparse recovery methods. These methods are Mip-NeRF \cite{barron2021mip}, DietNeRF \cite{jain2021putting} Reg-NeRF \cite{niemeyer2022regnerf}, FreeNeRF \cite{yang2023freenerf}, SparseNeRF \cite{wang2023sparsenerf}, 3DGS \cite{kerbl20233dsplatting}, DNGaussian \cite{li2024dngaussian}, FSGS \cite{zhu2024fsgs} and FewViewGS \cite{yin2024fewviewgs}. These methods are distributed between $3$ categories, the first category are methods that do not use any form of pre-trained methods, and these methods are Mip-NeRF, 3DGS and FreeNeRF. The second category are methods that use pre-trained methods that are not trained with 3D-aware data, such as image feature extractors, and these methods are DietNeRF and FewViewGs. The third category are methods that use 3D aware pre-trained methods such as depth predictors or methods trained on 3D data, and these methods are RegNeRF, SparseNeRF, DNGaussian and FSGS. DIP-GS belongs to the first category since it only relies on the structural regularization that DIP provides.

The quantitative results are presented in Table\ref{tab:blender_llff_resutls}.
On the \emph{Blender dataset} (8 input views), DIP-GS achieves state-of-the-art performance across all metrics, with a PSNR of $25.90$, SSIM of $0.898$, and LPIPS of $0.087$. Notably, it outperforms methods from all categories, including those relying on pre-trained models like FewViewGS* (PSNR 25.550) and FSGS** (PSNR 24.64). This demonstrates the strong regularization capability of DIP-GS in complex synthetic scenes.

On the \emph{LLFF dataset} (3 input views), which features real-world scenes and is generally more challenging for sparse-view reconstruction, DIP-GS remains highly competitive. It achieves a SOTA competitive PSNR of $20.13$ with a minor difference, and outperforms other methods with SSIM of $0.662$ and LPIPS of $0.221$. More importantly, DIP-GS significantly outperforms other pre-training-free methods like FreeNeRF (PSNR 19.63, SSIM 0.612, LPIPS 0.308) and vanilla 3DGS (PSNR 15.52) with a margin. 
In the \emph{DTU dataset} (3 input views), we obtain similar behaviour in which our method is SOTA competitive and outperforms it in many cases, as demonstrated in Table \ref{tab:blender_llff_resutls} and Figure \ref{fig:qualitive_results_llff}. Overall, DIP-GS demonstrates SOTA competitive capabilities and results.

Table \ref{tab:llff_dtu_6_9_views} presents the quantitative results on the LLFF and DTU datasets with $6$ and $9$ views respectively. Similar to the $3$ views case, our method manages to obtain SOTA results in most cases, specifically outperforms the pre-training free methods in these cases. These results further demonstrates the robustness and superiority of DIP-GS method. Because of the Unet activation at each iteration, DIP-GS method comes with a runtime overhead, in which each stage in it (DIP fitting + post-processing) takes approximately $20$ minutes to run on a A5000 gpu. Thus, in case of runtime limitation, one may sacrifice the performance little bit for the sake of runtime by letting DIP-GS run for fewer stages, or even one.

\section{Conclusion}
\label{sec:conclusions}

In this paper, we proposed DIP-GS, a method that generates Gaussians using DIP network, enabling a strongly regularized structure without pre-trained models. Our approach achieves competitive results on sparse-view datasets thanks to its regularized and structured prior. However, DIP-GS has some limitations: it incurs higher training time compared to standard GS-based methods due to the need to run a neural network at each optimization step, and it currently relies on a separate 3DGS post-processing stage for Gaussian densification. We leave integrating densification step directly into the DIP framework for future work, which could improve both efficiency and reconstruction quality. More broadly, DIP-GS paves the way for incorporating other DIP-based techniques, and single-image-based methods in general, into 3DGS, potentially enabling tasks like super-resolution, inpainting, and foreground-background separation.

\bibliography{main}

\clearpage

\twocolumn[
    \begin{center}
        {\LARGE \bf DIP-GS: Deep Image Prior For Gaussian Splatting Sparse View Recovery \\  Supplementary Materials}
    \end{center}
]

\maketitle

\section{Technical Details}
DIP-GS regularization weights are attached in Table \ref{tab:technical_details}. It is worth mentioning that scale and occlusion regularizations are only used with DTU dataset. For the mean and scale initialization stage, we use Adam optimizer with $lr=5e^{-3},1e^{-3}$ respectively, which we run each for $3000$ steps as mentioned in the paper. Before running the initialization process, all input Gaussians with opacity lower than $0.005$ are discarded. For the DIP optimization, we use AdamW optimizer with weight decay of $1e-5$ and $lr=2e^{-4},1e^{-3},1e^{-3},1e^{-3},1e^{-3}$ for $(f_{\theta_\mu}^{\mu}(\bm{z}),f_{\theta_o}^{o}(\bm{z}),f_{\theta_s}^{s}(\bm{z}),f_{\theta_r}^{r}(\bm{z}),f_{\theta_{sh}}^{sh}(\bm{z}))$ respectively. The DIP optimization runs for $4000$ steps as mentioned in the paper.

\begin{table}[ht]
\centering
\resizebox{0.48\textwidth}{!}{
    \begin{tabular}{ p{0.075\textwidth} | p{0.06\textwidth}|  p{0.06\textwidth} |p{0.06\textwidth}
    |p{0.06\textwidth}  |p{0.06\textwidth} |p{0.06\textwidth}  
    |p{0.06\textwidth}  |p{0.06\textwidth} |p{0.06\textwidth}|
    }

      &  $\beta_{init}$  & $\gamma_{init}$ & $\delta_{init}$ &  $\beta$ & $\gamma$ & $\delta$ &  $\beta_{post}$ & $\gamma_{post}$ & $\delta_{post}$ \\
     \hline
     
     Blender &  $0.05$ & $0$ & $0$ &   $0.02$ & $0$ & $0$ & $0.02$ & $0$ & $0$\\
     \hline

     LLFF &  $0.1$ & $0$ & $0$ &   $0.02$ & $0$ & $0$ & $0.05$ & $0$ & $0$\\
     \hline

     DTU &  $0.1$ & $0.1$ & $20$ &   $0.02$ & $0.01$ & $20$ & $0.05$ & $0.01$ & $20$\\
     \hline
     
    \end{tabular}
}
    \caption{Hyperparameters values for the different datasets. $\beta,\gamma$ and $\delta$ are the weights for opacity, scale and occlusion regularization respectively.
    Init for the initial stage, middle is for the DIP optimization, and post for the post-processing.}
    \label{tab:technical_details}
     \vspace{-0.2in}
\end{table}

\section{Qualitative Results}
\begin{figure*}[h]
\centering
    \begin{tabular}{   p{0.17\textwidth}  p{0.17\textwidth} p{0.17\textwidth}  p{0.17\textwidth} p{0.17\textwidth}}

    \includegraphics[width=0.19\textwidth]{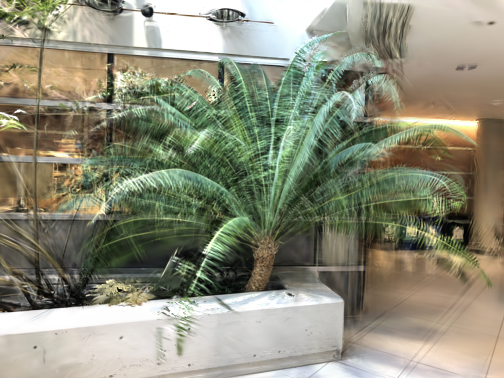} &
    \includegraphics[width=0.19\textwidth]{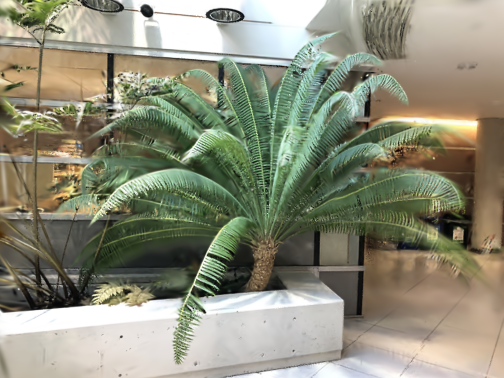} &
    \includegraphics[width=0.19\textwidth]{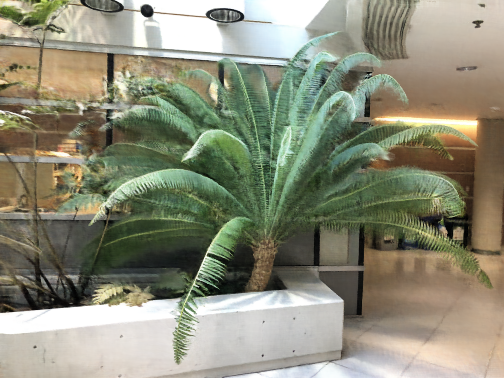} &    \includegraphics[width=0.19\textwidth]{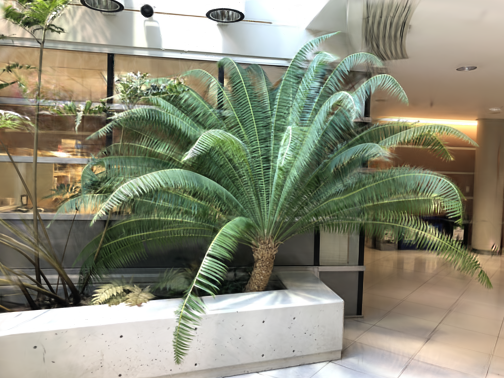} &
    \includegraphics[width=0.19\textwidth]{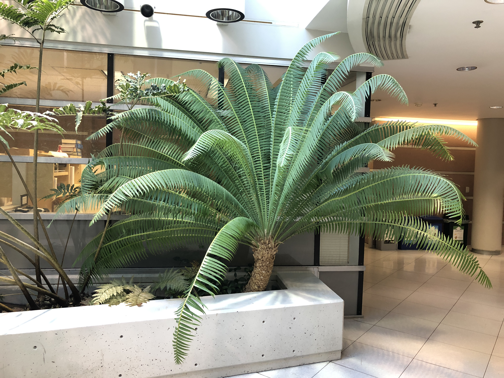}  \\

    \includegraphics[width=0.19\textwidth]{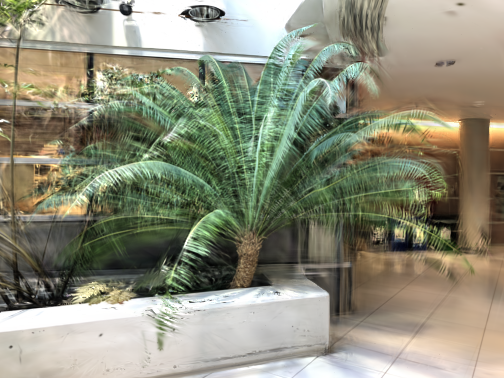} &
    \includegraphics[width=0.19\textwidth]{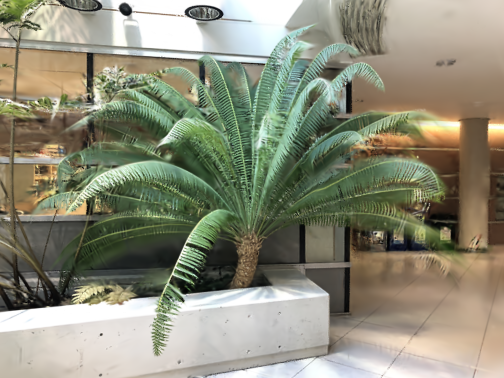} &
    \includegraphics[width=0.19\textwidth]{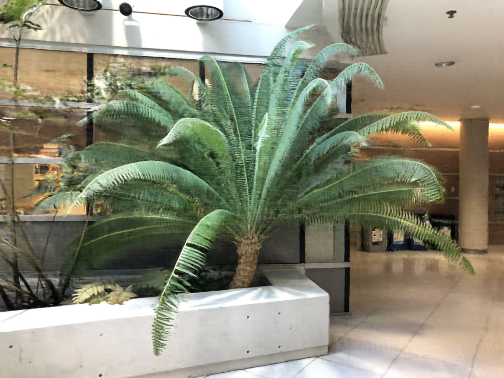} &    \includegraphics[width=0.19\textwidth]{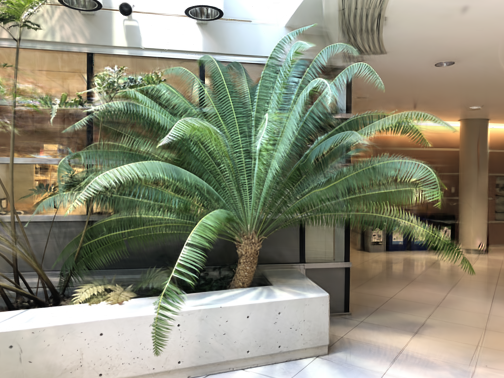} &
    \includegraphics[width=0.19\textwidth]{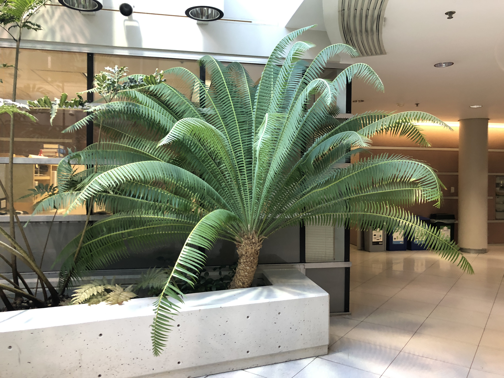}  \\

    \includegraphics[width=0.19\textwidth]{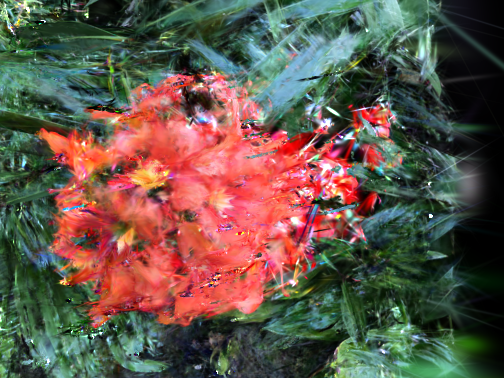} &
    \includegraphics[width=0.19\textwidth]{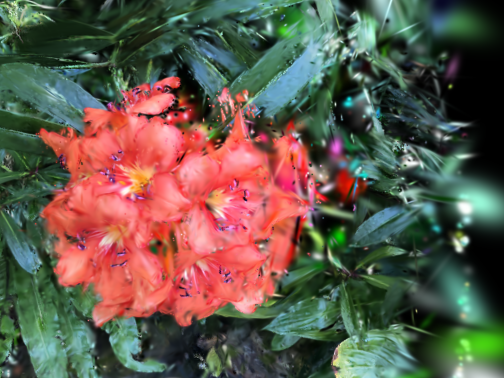} &
    \includegraphics[width=0.19\textwidth]{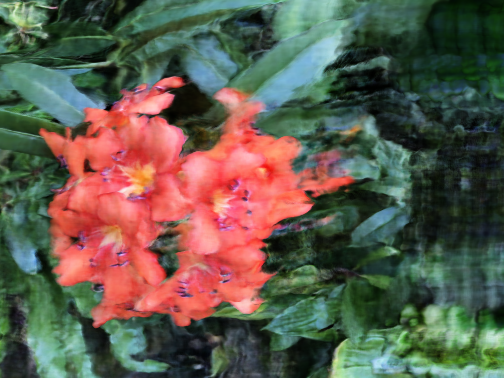} &    \includegraphics[width=0.19\textwidth]{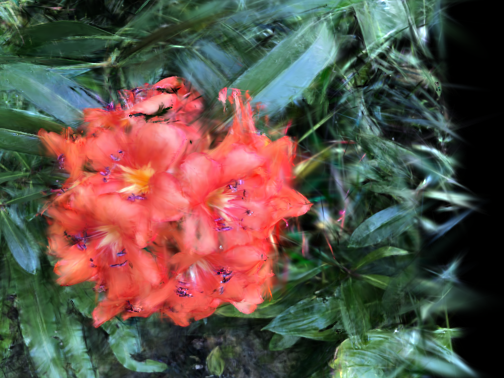} &
    \includegraphics[width=0.19\textwidth]{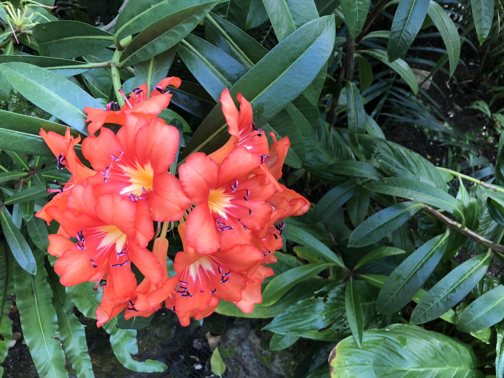}  \\

    \includegraphics[width=0.19\textwidth]{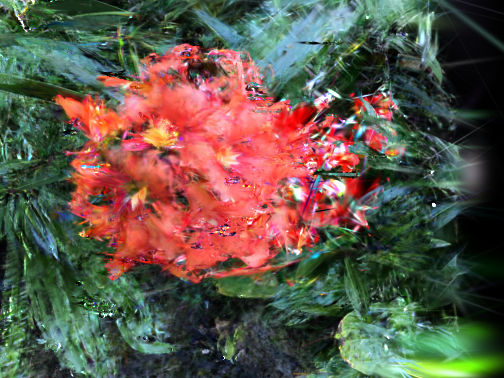} &
    \includegraphics[width=0.19\textwidth]{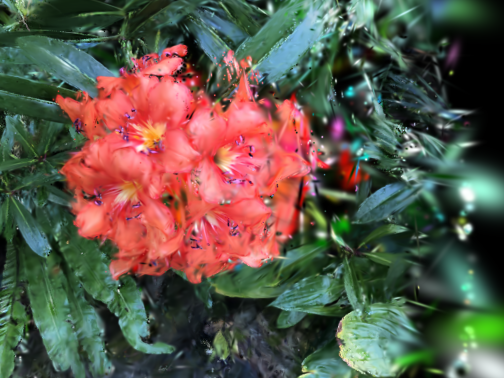} &
    \includegraphics[width=0.19\textwidth]{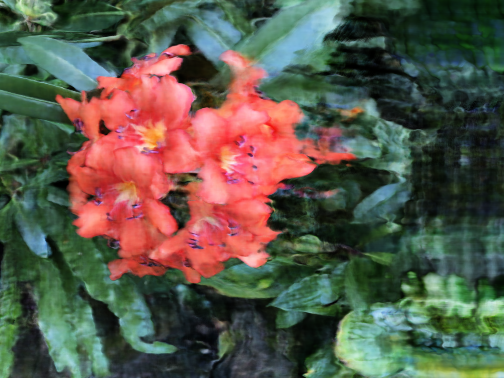} &    \includegraphics[width=0.19\textwidth]{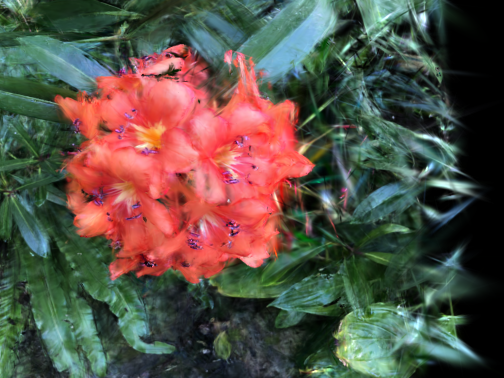} &
    \includegraphics[width=0.19\textwidth]{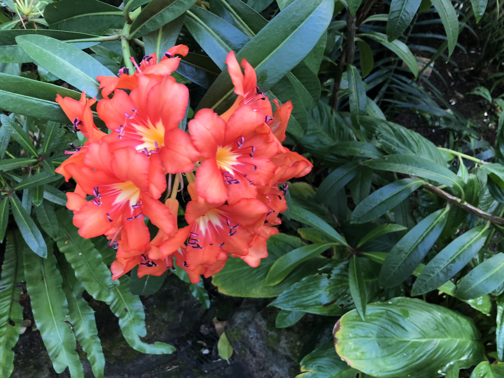}  \\

    \includegraphics[width=0.19\textwidth]{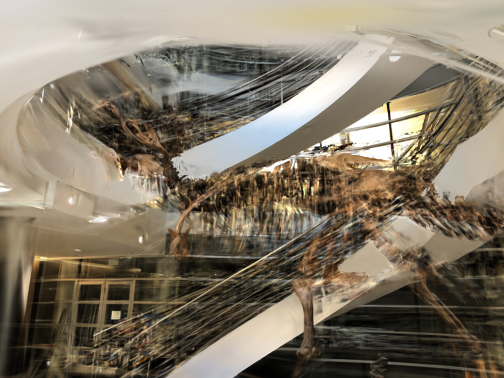} &
    \includegraphics[width=0.19\textwidth]{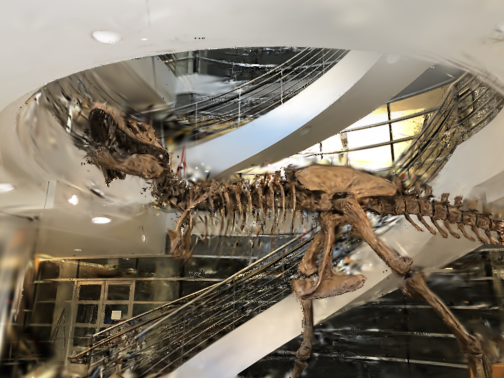} &
    \includegraphics[width=0.19\textwidth]{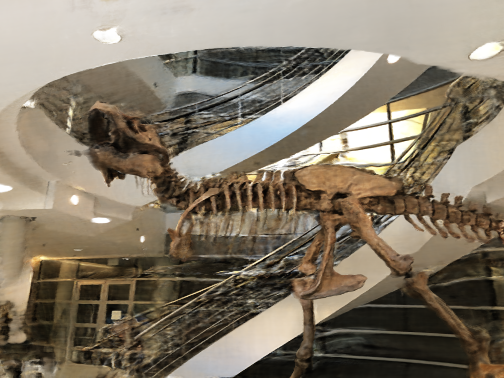} &    \includegraphics[width=0.19\textwidth]{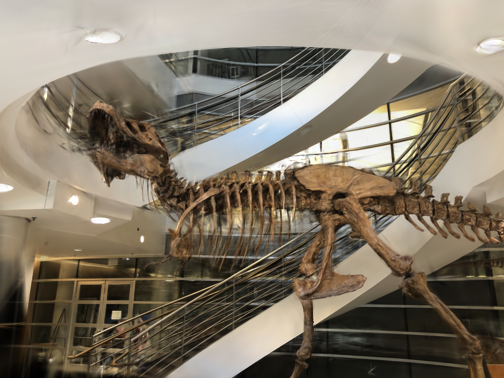} &
    \includegraphics[width=0.19\textwidth]{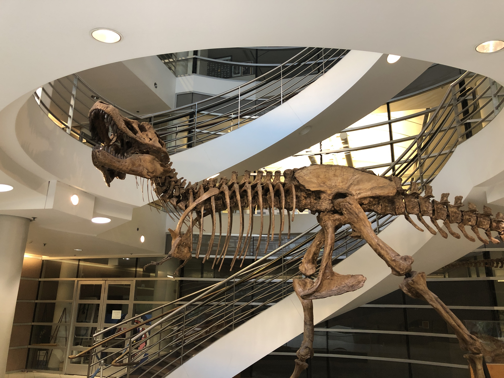}  \\

    \includegraphics[width=0.19\textwidth]{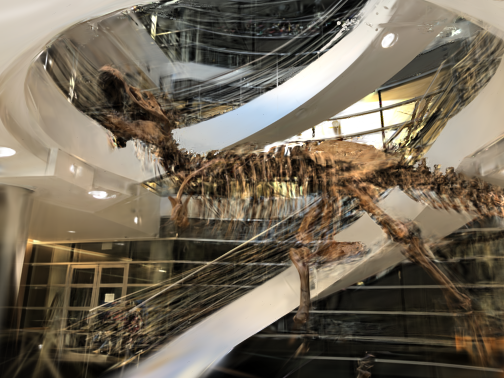} &
    \includegraphics[width=0.19\textwidth]{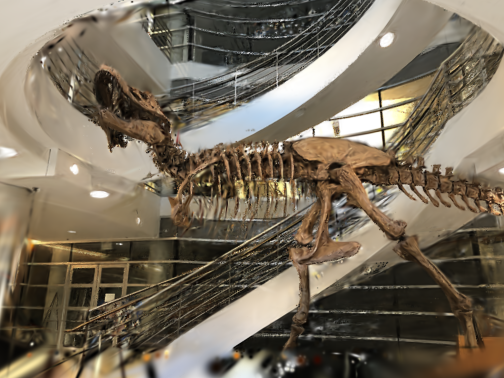} &
    \includegraphics[width=0.19\textwidth]{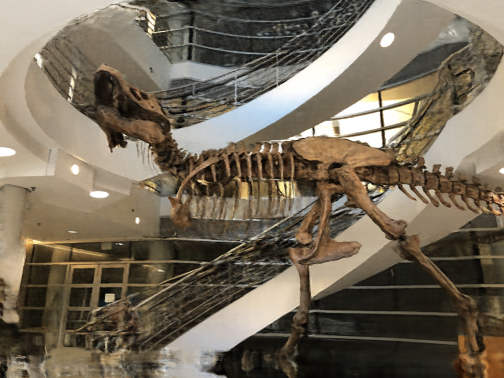} &    \includegraphics[width=0.19\textwidth]{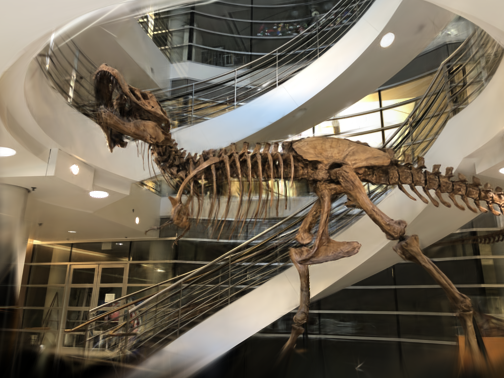} &
    \includegraphics[width=0.19\textwidth]{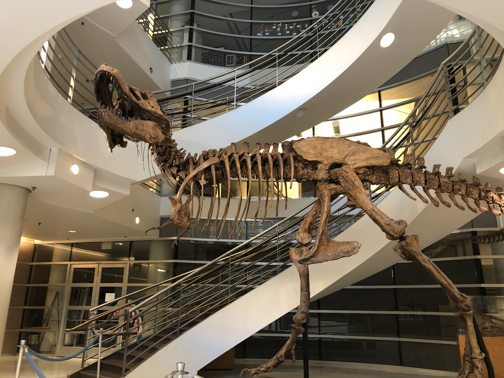}  \\

    \includegraphics[width=0.19\textwidth]{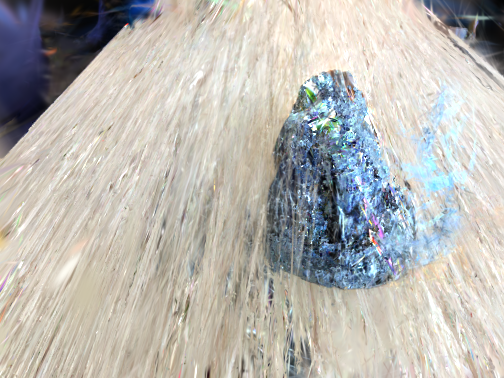} &
    \includegraphics[width=0.19\textwidth]{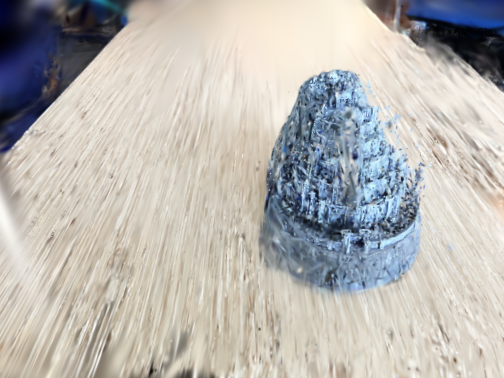} &
    \includegraphics[width=0.19\textwidth]{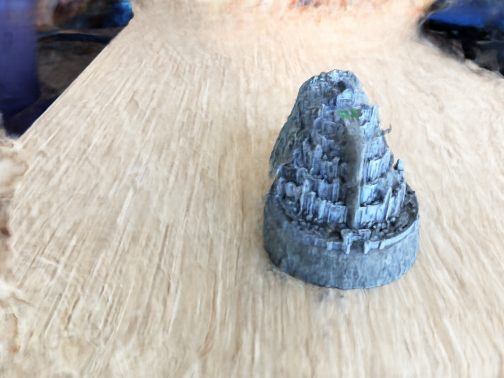} &    \includegraphics[width=0.19\textwidth]{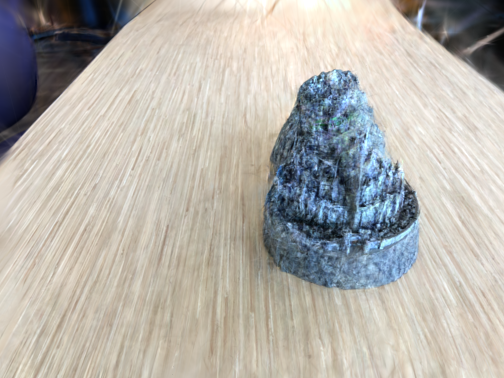} &
    \includegraphics[width=0.19\textwidth]{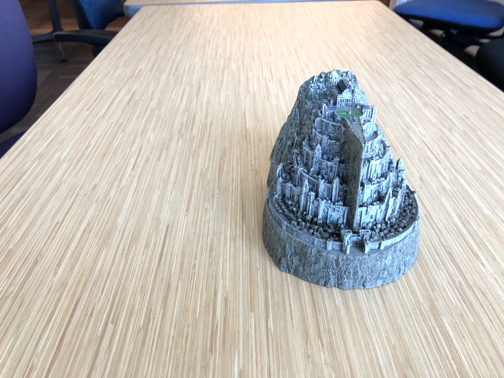}  \\

    \includegraphics[width=0.19\textwidth]{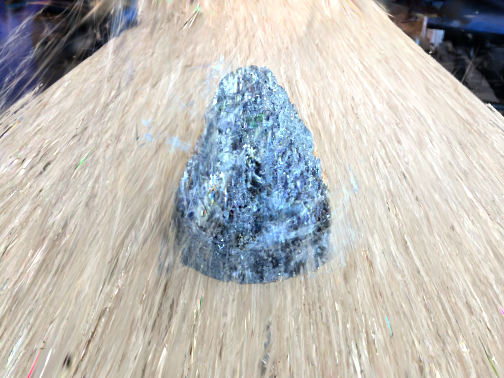} &
    \includegraphics[width=0.19\textwidth]{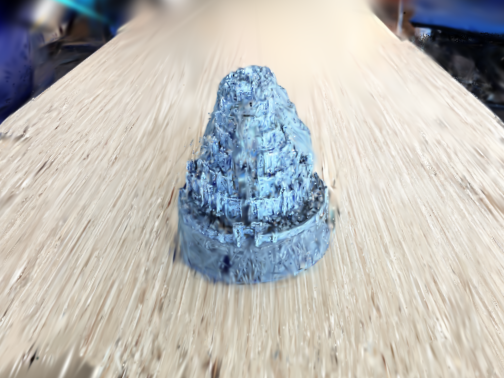} &
    \includegraphics[width=0.19\textwidth]{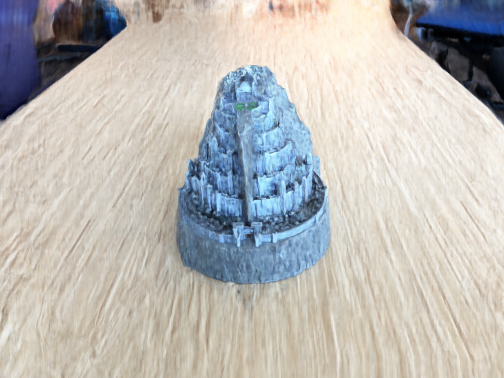} &    \includegraphics[width=0.19\textwidth]{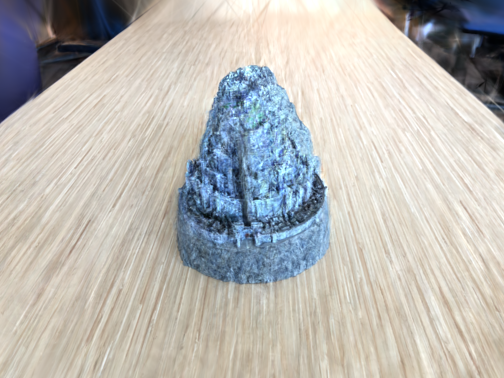} &
    \includegraphics[width=0.19\textwidth]{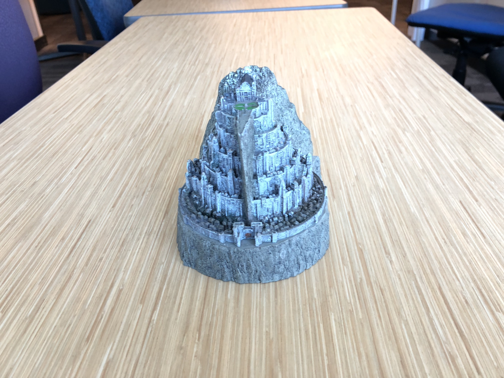}  \\

    \multicolumn{1}{c}{3DGS} & \multicolumn{1}{c}{DNGaussian}  & \multicolumn{1}{c}{FreeNeRF} &\multicolumn{1}{c}{DIP-GS} & \multicolumn{1}{c}{GT}

    \end{tabular}
    \caption{LLFF qualitative results}
    \label{fig:qualitive_results_supp_LLFF}
\end{figure*}

\begin{figure*}[t]
\centering
    \begin{tabular}{   p{0.32\textwidth}  p{0.32\textwidth} p{0.32\textwidth} }

    \includegraphics[width=0.31\textwidth]{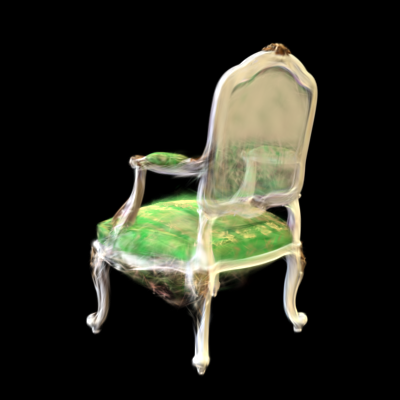} &
    \includegraphics[width=0.31\textwidth]{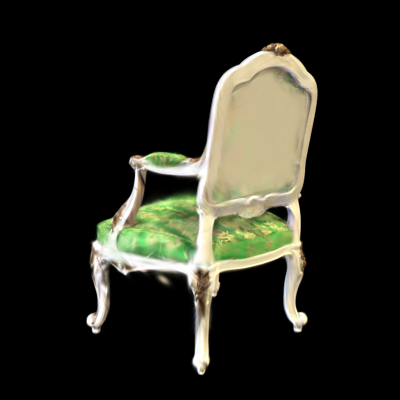} &
    \includegraphics[width=0.31\textwidth]{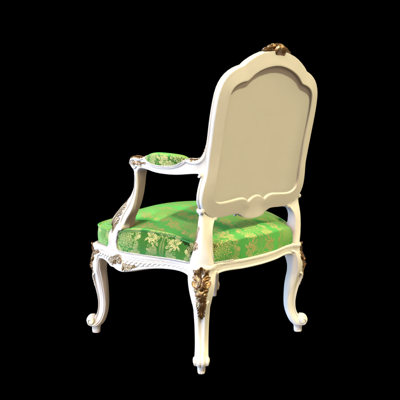} \\

    \includegraphics[width=0.31\textwidth]{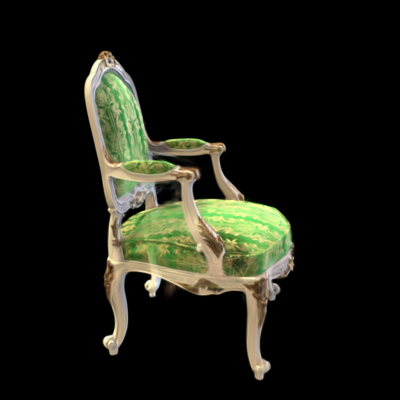} &
    \includegraphics[width=0.31\textwidth]{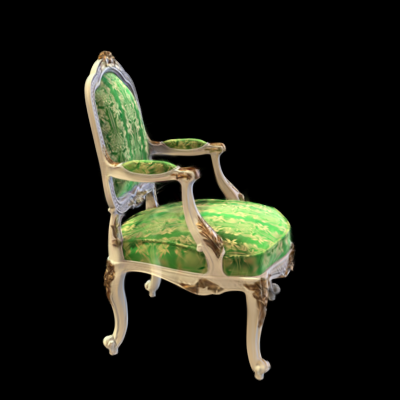} &
    \includegraphics[width=0.31\textwidth]{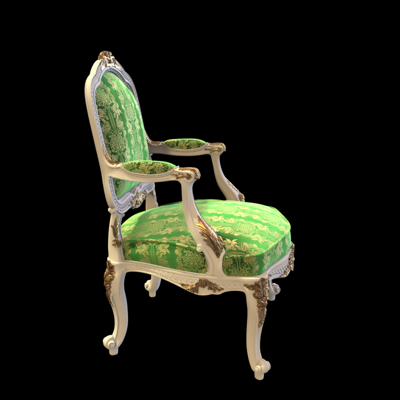} \\

    \includegraphics[width=0.31\textwidth]{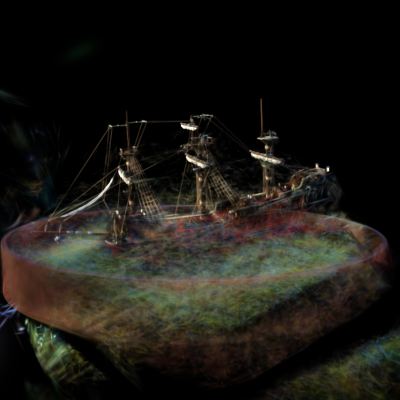} &
    \includegraphics[width=0.31\textwidth]{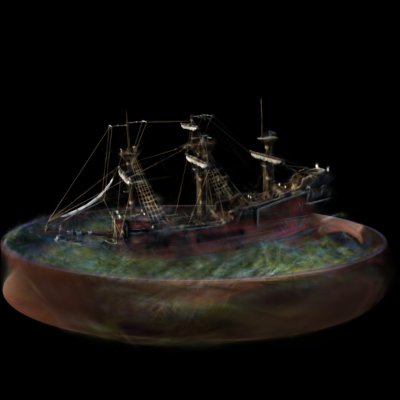} &
    \includegraphics[width=0.31\textwidth]{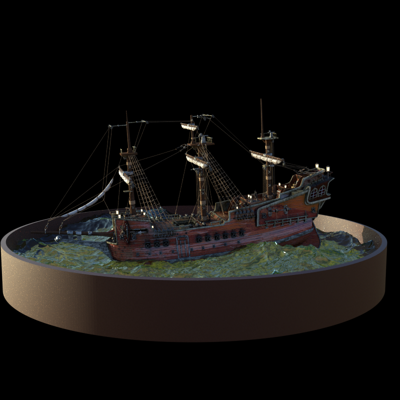} \\

    \includegraphics[width=0.31\textwidth]{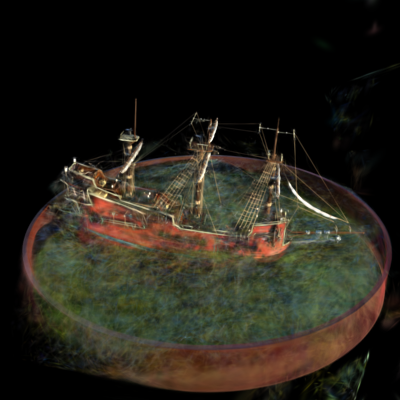} &
    \includegraphics[width=0.31\textwidth]{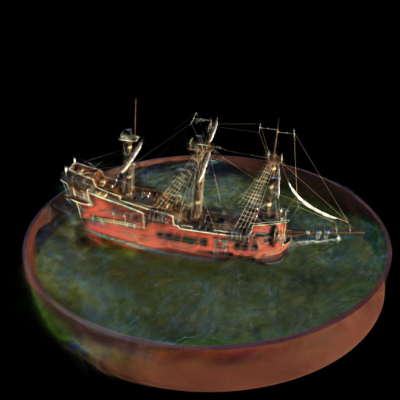} &
    \includegraphics[width=0.31\textwidth]{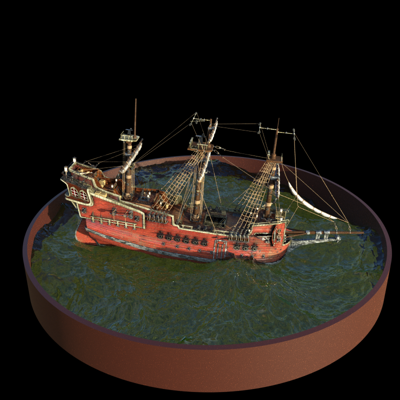} \\

    \multicolumn{1}{c}{3DGS}   &\multicolumn{1}{c}{DIP-GS} & \multicolumn{1}{c}{GT}

    \end{tabular}
    \caption{Blender qualitative results.}
    \label{fig:qualitive_results_supp_blender}
\end{figure*}

\begin{figure*}[h]
\centering
    \begin{tabular}{   p{0.17\textwidth}  p{0.17\textwidth} p{0.17\textwidth}  p{0.17\textwidth} p{0.17\textwidth}}

    \includegraphics[width=0.19\textwidth]{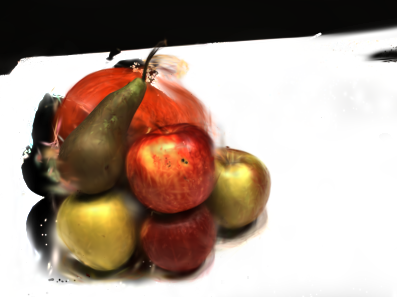} &
    \includegraphics[width=0.19\textwidth]{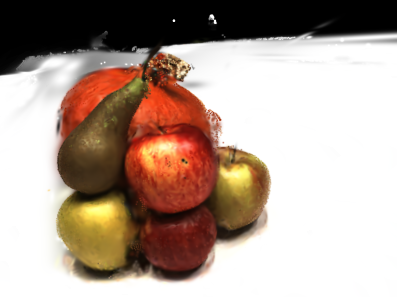} &
    \includegraphics[width=0.19\textwidth]{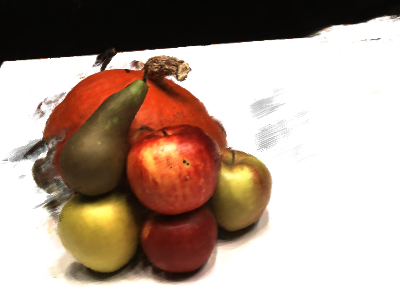} &    \includegraphics[width=0.19\textwidth]{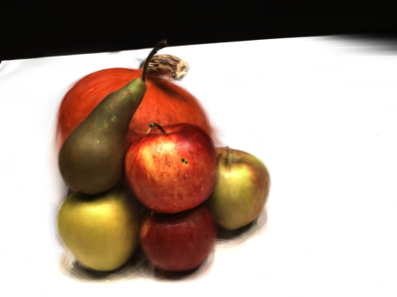} &
    \includegraphics[width=0.19\textwidth]{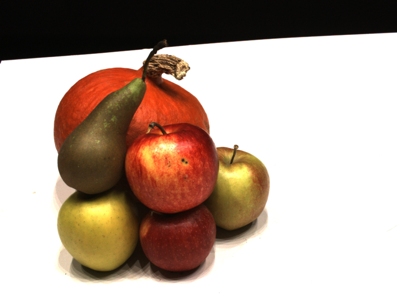}  \\

    \includegraphics[width=0.19\textwidth]{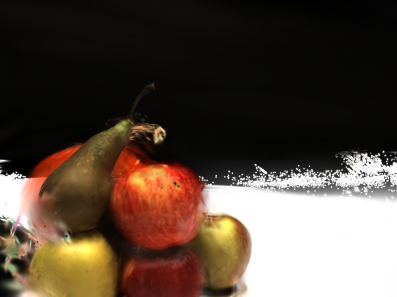} &
    \includegraphics[width=0.19\textwidth]{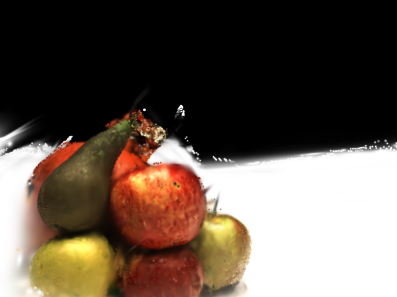} &
    \includegraphics[width=0.19\textwidth]{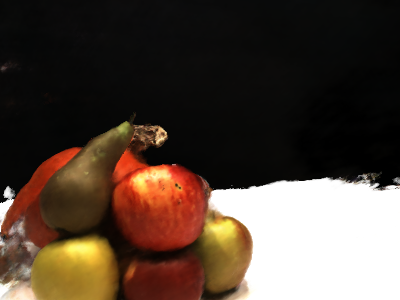} &    \includegraphics[width=0.19\textwidth]{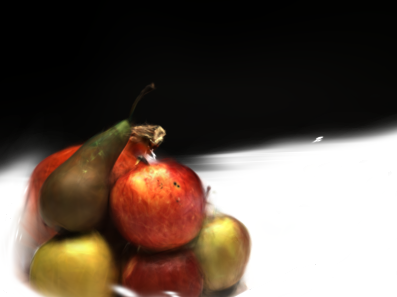} &
    \includegraphics[width=0.19\textwidth]{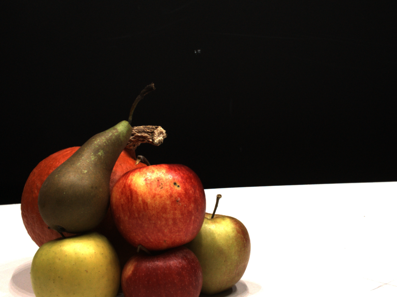}  \\

    \includegraphics[width=0.19\textwidth]{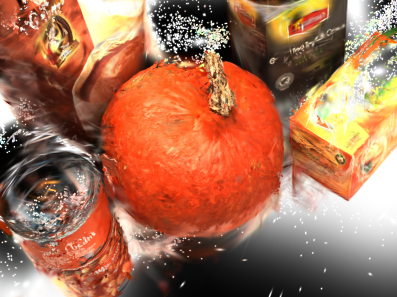} &
    \includegraphics[width=0.19\textwidth]{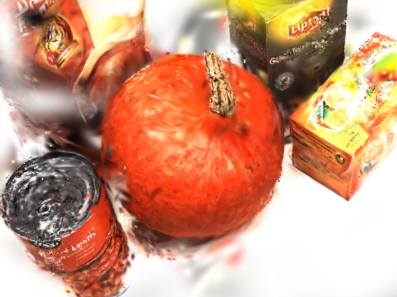} &
    \includegraphics[width=0.19\textwidth]{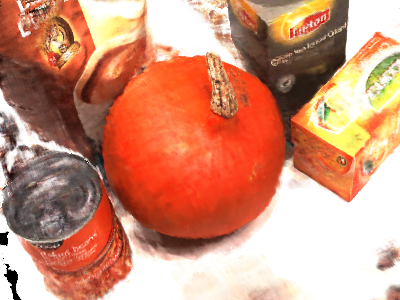} &    \includegraphics[width=0.19\textwidth]{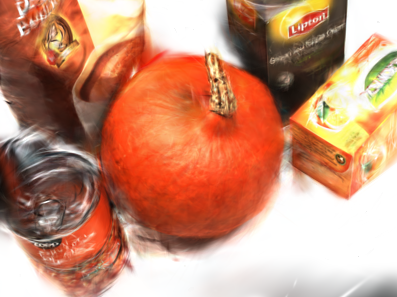} &
    \includegraphics[width=0.19\textwidth]{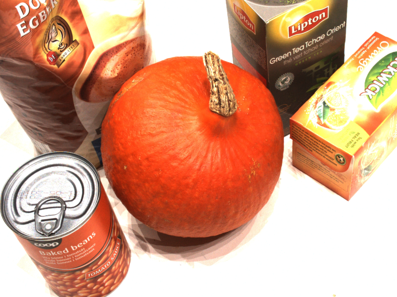}  \\

    \includegraphics[width=0.19\textwidth]{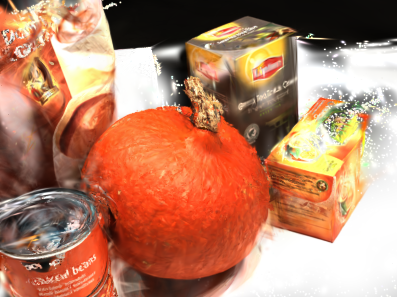} &
    \includegraphics[width=0.19\textwidth]{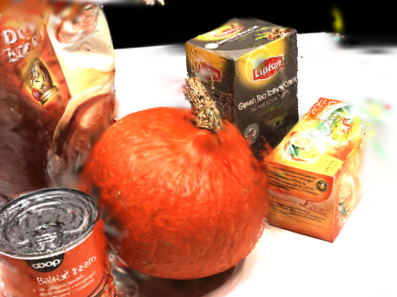} &
    \includegraphics[width=0.19\textwidth]{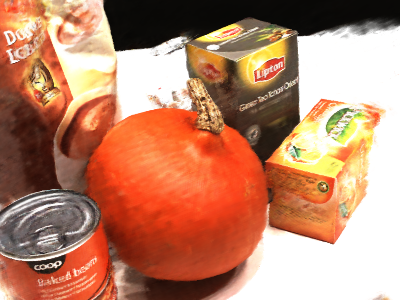} &    \includegraphics[width=0.19\textwidth]{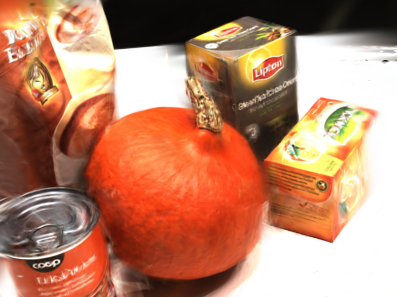} &
    \includegraphics[width=0.19\textwidth]{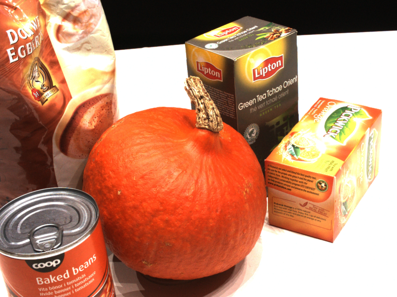}  \\

    \includegraphics[width=0.19\textwidth]{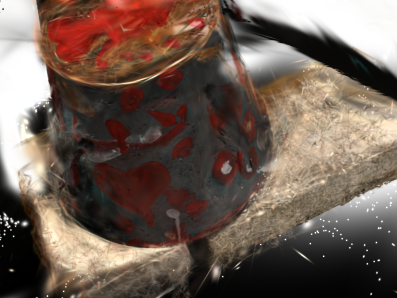} &
    \includegraphics[width=0.19\textwidth]{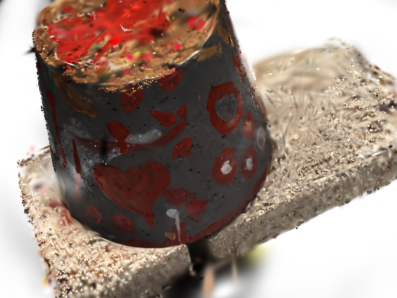} &
    \includegraphics[width=0.19\textwidth]{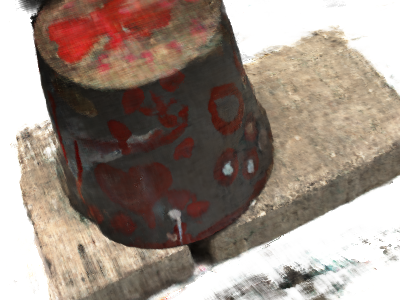} &    \includegraphics[width=0.19\textwidth]{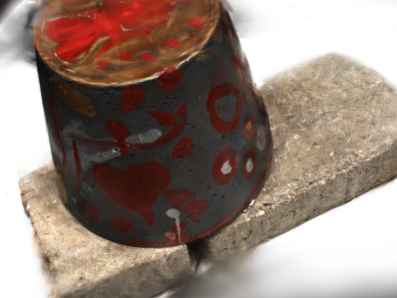} &
    \includegraphics[width=0.19\textwidth]{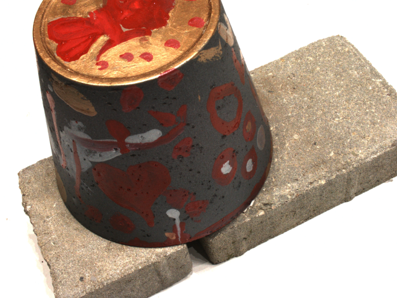}  \\

    \includegraphics[width=0.19\textwidth]{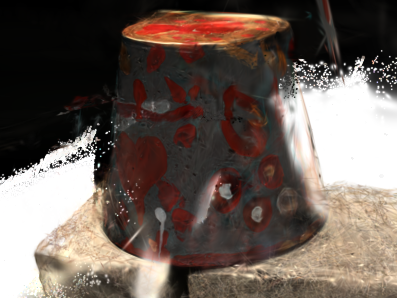} &
    \includegraphics[width=0.19\textwidth]{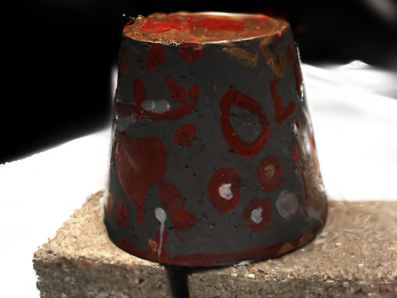} &
    \includegraphics[width=0.19\textwidth]{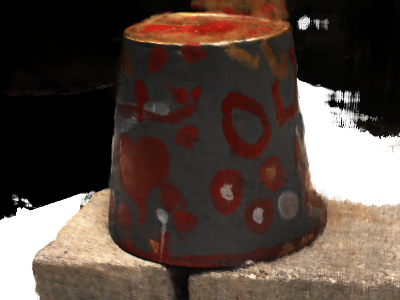} &    \includegraphics[width=0.19\textwidth]{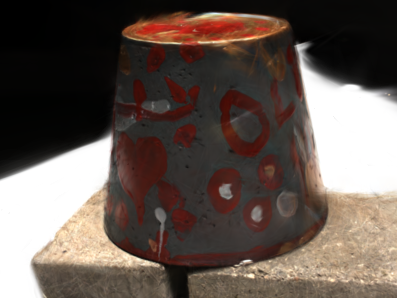} &
    \includegraphics[width=0.19\textwidth]{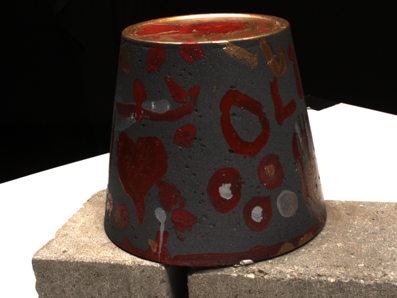}  \\

    \includegraphics[width=0.19\textwidth]{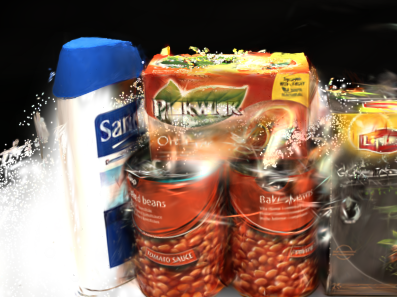} &
    \includegraphics[width=0.19\textwidth]{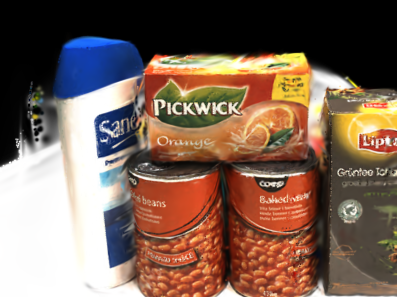} &
    \includegraphics[width=0.19\textwidth]{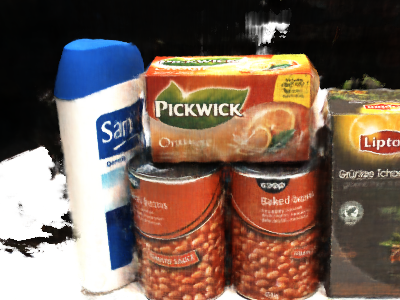} &    \includegraphics[width=0.19\textwidth]{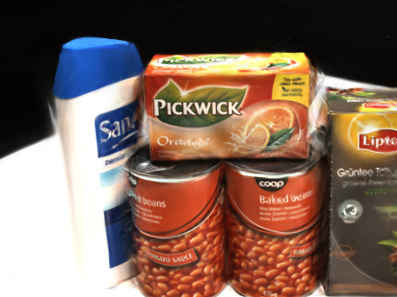} &
    \includegraphics[width=0.19\textwidth]{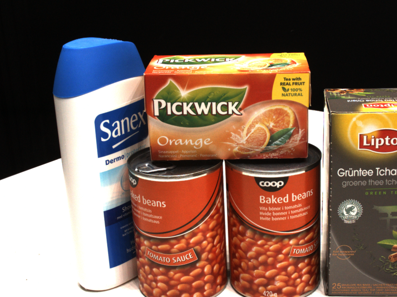}  \\

    \includegraphics[width=0.19\textwidth]{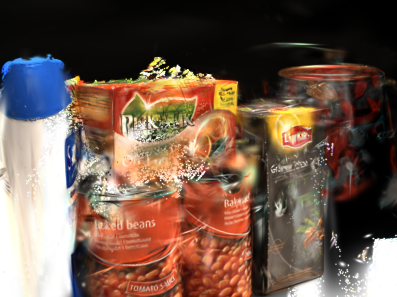} &
    \includegraphics[width=0.19\textwidth]{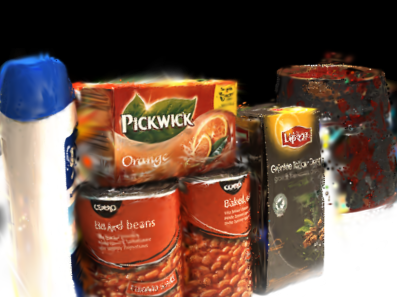} &
    \includegraphics[width=0.19\textwidth]{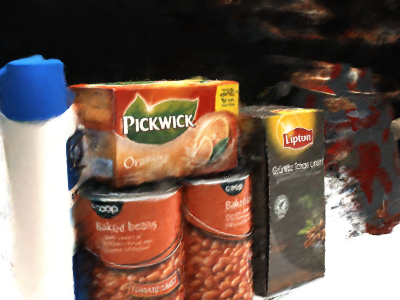} &    \includegraphics[width=0.19\textwidth]{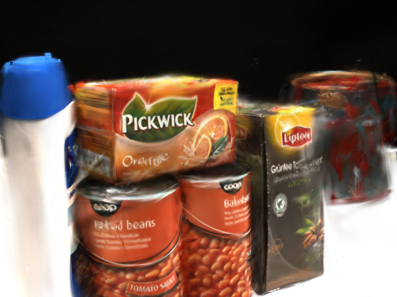} &
    \includegraphics[width=0.19\textwidth]{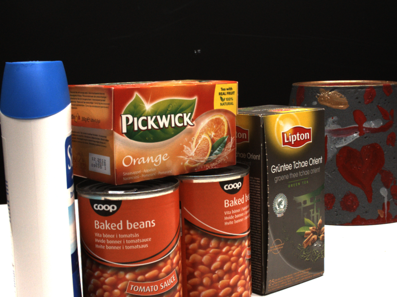}  \\

    \multicolumn{1}{c}{3DGS} & \multicolumn{1}{c}{DNGaussian}  & \multicolumn{1}{c}{FreeNeRF} &\multicolumn{1}{c}{DIP-GS} & \multicolumn{1}{c}{GT}

    \end{tabular}
    \caption{DTU qualitative results}
    \label{fig:qualitive_results_supp_DTU}
\end{figure*}

Figures \ref{fig:qualitive_results_supp_LLFF}, \ref{fig:qualitive_results_supp_blender}, \ref{fig:qualitive_results_supp_DTU} include additional qualitative results for DIP-GS method.

\end{document}